\documentclass[sigconf]{acmart}

\copyrightyear{2026}
\acmYear{2026}
\setcopyright{cc}
\setcctype{by}
\acmConference[SIGIR '26] {Proceedings of the 49th International ACM SIGIR Conference on Research and Development in Information Retrieval}{July 20--24, 2026}{Melbourne, VIC, Australia.}
\acmBooktitle{Proceedings of the 49th International ACM SIGIR Conference on Research and Development in Information Retrieval (SIGIR '26), July 20--24, 2026, Melbourne, VIC, Australia}
\acmISBN{979-8-4007-2599-9/2026/07}
\acmDOI{10.1145/3805712.3809547}
\usepackage{enumitem}
\usepackage{booktabs} 
\usepackage{amsmath}  
\usepackage[utf8]{inputenc} 
\usepackage{textcomp}
\usepackage{graphicx}
\usepackage{subcaption}
\usepackage{multirow}
\usepackage{hyperref}
\usepackage{balance}
\begin{document}

\title{Debiased Multimodal Personality Understanding through Dual Causal Intervention}

\author{Yangfu Zhu}
\affiliation{%
  \institution{Capital Normal University}
  \city{Beijing}
  \country{China}
}
\email{zhuyangfu@cnu.edu.cn}

\author{Zitong Han}
\affiliation{%
  \institution{Capital Normal University}
  \city{Beijing}
  \country{China}
}
\email{hanzitong@cnu.edu.cn}

\author{Nianwen Ning}
\affiliation{%
  \institution{Henan University}
  \city{Henan}
  \country{China}
}
\email{nnw@henu.edu.cn}

\author{Yuting Wei}
\affiliation{%
  \institution{University of International Relations}
  \city{Beijing}
  \country{China}
}
\email{weiyuting@uir.edu.cn}

\author{Yuandong Wang}
\affiliation{%
  \institution{Capital Normal University}
  \city{Beijing}
  \country{China}
}
\email{wangyd@cnu.edu.cn}

\author{Hang Feng}
\affiliation{%
  \institution{Capital Normal University}
  \city{Beijing}
  \country{China}
}
\email{fenghang@cnu.edu.cn}

\author{Zhenzhou Shao}
\authornote{Corresponding author}
\affiliation{%
  \institution{Capital Normal University}
  \city{Beijing}
  \country{China}
}
\email{zshao@cnu.edu.cn}

\renewcommand{\shortauthors}{Yangfu Zhu et al.}

\begin{abstract}
Multimodal personality understanding plays a critical role in human-centered artificial intelligence. Previous work mainly focus on learning rich multimodal representations for video personality understanding. However, they often suffer from potential harm caused by subject bias (e.g., observable age and unobservable mental states), as subjects originate from diverse demographic backgrounds. Learning such spurious associations between multimodal features and traits may lead to unfair personality understanding. In this work, we construct a Structural Causal Model (SCM) to analyze the impact of these biases from a causal perspective, and propose a novel \textbf{D}ual \textbf{C}ausal \textbf{A}djustment Network \textbf{(DCAN)} to mitigate the interference of subject attributes on personality understanding. 
Specifically,  we design a Back-door Adjustment Causal Learning (BACL) module to block spurious correlations from observable demographic factors via a prototype-based confounder dictionary, and subsequently apply a Front-door Adjustment Causal Learning (FACL) module to address latent and unobservable biases through a learned mediator dictionary intervention, thereby achieving causal disentanglement of representations for deconfounded reasoning. 
Importantly, we construct a \textbf{D}emographic-annotated \textbf{M}ultimodal \textbf{S}tudent \textbf{P}ersonality \textbf{(DMSP)} dataset to support the analysis and discussion of fairness-related factors.
Extensive experiments on the benchmark dataset CFI-V2 and our DMSP dataset demonstrate that DCAN consistently improves prediction accuracy, reaching 92.11\% and 92.90\%, respectively.
Meanwhile, the improvements in the fairness metrics of equal opportunity and demographic parity are 6.57\% and 7.97\% on CFI-V2, and 15.38\% and 20.06\% on the DMSP dataset. Our code and DMSP dataset are available
at \url{https://github.com/Sabrina-han/DCAN}
\end{abstract}





\maketitle
\section{Introduction}

With the popularity of video social platforms such as TikTok and YouTube, people have grown increasingly accustomed to expressing their inner thoughts and recording their feelings through the multimodal video format \cite{zhao2025attention}. The multimodal personality understanding task aims to infer intrinsic personality traits such as the Big Five \cite{mccrae1992introduction} traits and the Myers-Briggs Type Indicator (MBTI) from visual, auditory, and linguistic signals in videos \cite{leekha2024vyaktitvanirdharan,liao2024open}. Unlike subjective self-reports, such data-driven methods offer objective and scalable solutions for applications across human-computer interaction \cite{song2025predicting}, mental health monitoring \cite{yin2025mdd}, and personalized recommendation \cite{wang2025towards}. Thus, accurate personality understanding from multimodal data is crucial for advancing user-centric AI systems.


Previous work mainly focused on visual cues in videos~\cite{biel2013youtube, gurpinar2016transfer}, attempting to map non-verbal signals directly to personality traits. However, relying on a single modality results in information sparsity and increases susceptibility to noise.
Recent studies have increasingly focused on multimodal fusion methods\cite{majumder2017deep, kampman2018investigating, sun2018personality, jiang2020multimodal}. Specifically, some approaches treat language as the primary modality and integrate acoustic and visual cues, leveraging multimodal attention mechanisms or joint representation learning to improve personality prediction performance\cite{li2020crnet,suman2022multi}.  Some studies further explore fine-grained cross-modal interaction modeling strategies to enhance the expressiveness and robustness of multimodal personality representations \cite{masumura2025multimodal,yang2021graph}. Despite achieving high accuracy through the fusion of heterogeneous modalities, existing models are often susceptible to subject bias, undermining the fairness and reliability of personality understanding.

\begin{figure}[t]
  \centering
  \includegraphics[width=\linewidth]{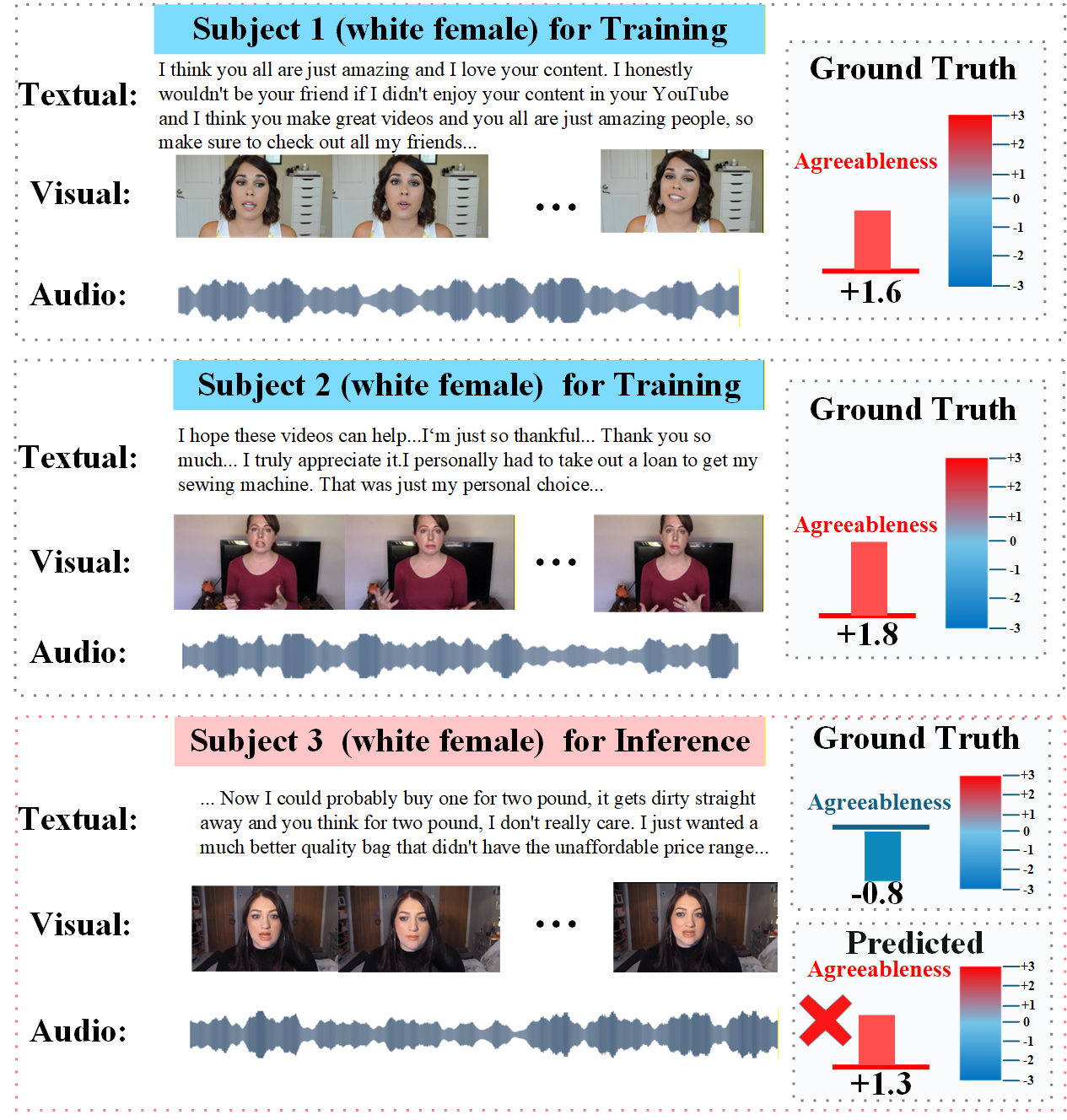}
  \caption{A case illustrates how subject confounders introduce spurious correlations in personality understanding. Subjects 1 and 2 (Training) associate white female appearances and gentle tones with high agreeableness, misleading the model to infer  Subject 3 (white female with low agreeableness).}
  \label{fig:bias_illustration}
\end{figure}


From the causal perspective, subject-related demographic attributes and unobservable factors are essentially regarded as confounders, which mislead the models to learn spurious correlations between user-generated multimodal data and traits in the training data, as well as causing prediction bias for new subjects in the inference. 
The examples in Figure~\ref{fig:bias_illustration} provide strong evidence of this subject bias. Concretely, subjects 1 and 2 in the training set share a specific demographic attribute (i.e., white female) and consistently exhibit high agreeableness scores. In this case, the trained model is misled to focus on spurious clues from subject-specific appearances rather than genuine behavioral semantics. It inexorably establishes a spurious connection between female characteristics (e.g., facial appearance or acoustic timbre) and high agreeableness levels. Consequently, the model erroneously takes these gender-related features as evidence for the personality trait and makes an entirely incorrect prediction when applied to subject 3 (white female with low agreeableness ), failing to generalize well across new subjects. In addition, unobservable confounding factors such as psychological state, emotional fluctuations, and social background are also important factors influencing model predictions. These confounding factors cannot be directly captured through traditional visual or acoustic features, but they may, in certain cases, implicitly lead the model to learn incorrect associations in the data.

However, in tackling the aforementioned issue, we face the following challenges:
\textbf{(1) }Latent confounders are difficult to measure directly, and how can these potential confounding factors be identified? \textbf{(2)} How to construct a unified de-biasing model to handle both observable and unobservable confounders?

To address the aforementioned challenges, we first establish a Structural Causal Model (SCM) to characterize the causal relationships underlying the multimodal personality understanding task.   Based on this formulation, we propose a novel \textbf{D}ual \textbf{C}ausal \textbf{A}djustment Network \textbf{(DCAN)} to mitigate the interference of subject attributes.  Unlike neural network mappings that rely on likelihood estimation $P(Y|X)$,  DCAN explicitly embraces causal intervention $P(Y|do(X))$ under the back-door and front-door adjustment to achieve deconfounded learning.
Specifically, we design a Back-door Adjustment Causal Learning (BACL) module to block spurious correlations induced by observable demographic confounders via a prototype-based confounder dictionary. Furthermore, a Front-door Adjustment Causal Learning (FACL) module is introduced to alleviate latent and unobservable biases through intervention on learned mediator dictionaries. Finally, de-biased multimodal representations are leveraged for personality understanding.
To the best of our knowledge, existing multimodal personality datasets with demographic annotations are primarily based on the Big Five traits. To further advance fairness research, we additionally construct a \textbf{D}emographic-annotated \textbf{M}ultimodal \textbf{S}tudent \textbf{P}ersonality \textbf{(DMSP)} dataset under the MBTI taxonomy, which enables systematic analysis and discussion of fairness-related factors.
To sum up, this paper has the following contributions:
\begin{itemize}
    \item We are the first to investigate subject bias in multimodal personality understanding and analyze its underlying causal mechanisms.
    
    \item The proposed DCAN end-to-end integrates back-door and front-door adjustment mechanisms to jointly mitigate the effects of observable demographic and latent confounders.

  \item   We construct a new Demographic-annotated Multimodal Student Personality (DMSP)  dataset under the MBTI taxonomy that supports in-depth analysis of subject bias and fairness in personality understanding, filling a gap in existing benchmarks.
    
    \item  Extensive experiments demonstrate that our DCAN framework significantly outperforms state-of-the-art baselines in both accuracy and fairness.
\end{itemize}

\section{Related Work}
\subsection{Multimodal Personality Understanding}
The Big Five personality traits provide a widely accepted framework for understanding individual differences \cite{digman1990personality}. While traditionally measured through resource-intensive self-report questionnaires, the rapid progress of affective computing has established multimodal personality recognition as a central task in human behavior analysis.
Early works predominantly rely on hand-crafted visual and acoustic features\cite{bell2008personality, schuller2011automatic, kautz2006extracting}. With deep learning, multimodal methods emerge to enhance personality prediction, including PersEmoN \cite{zhang2019persemon} for joint personality-emotion modeling, SENet \cite{hu2018multimodal} for cross-modal feature interactions, and CRNet \cite{li2020crnet} for audiovisual-textual fusion. 
Subsequent works introduce more sophisticated architectures: MM-ResVGG \cite{suman2022multi} integrates ResNet, VGGish, and n-gram CNN for three-modal fusion, while TBi-LSTM \cite{zhao2023integrating} combines transformers with Bi-LSTM for improved cross-modal alignment. More recently, OCEAN-AI \cite{ryumina2024ocean} leverages facial analysis via EmoFormer \cite{ryumina2024emoformer}, and PMGRT \cite{wang2025novel} introduces graph relational transformers to capture temporal dynamics. SSM-based approaches such as VideoMamba \cite{li2024videomamba} and Cobra \cite{zhao2025cobra} address efficiency bottlenecks for long-range modeling. In 2025, PINet \cite{tang2025pose} incorporates human pose for interpretable inference, FG-PTM \cite{masumura2025finegrained} proposes fine-grained item-level prediction, and then KE-HHG \cite{song2025graph} designs knowledge-enhanced hierarchical heterogeneous graphs for limited data scenarios.
However, despite high accuracy from heterogeneous modality fusion, existing models are still prone to subject bias. Such biases induce spurious correlations, which impair the fairness and reliability of personality assessment and hinder practical deployment.

\subsection{Multimodal Personality Datasets}
Existing multimodal personality datasets vary greatly in scope, scale, and annotation quality. 
Early works such as SIAP~\cite{giritlioglu2021siap}, which collects short interview videos annotated with Big Five scores, and Vyaktitv~\cite{sharma2022vyaktitv}, which focuses on Hindi-speaking subjects with manually rated traits, target limited participant groups and contain small-scale samples, restricting their generalizability. 
Larger benchmarks such as the ChaLearn First Impressions (CFI-V2)~\cite{ponce2016chalearn}, a collection of YouTube interview clips annotated for apparent personality traits, and UDIVA~\cite{udiva2021} extend to conversational dyads, providing rich multimodal cues from video, audio, and text. 
Recent studies shift toward fine-grained and domain-specific personality understanding. 
RecruitView ~\cite{recruitview2025} introduces comparative labels for candidate evaluation, Masumura et al.~\cite{masumura2025aaai} propose item-level annotations for subtle expressions, and CPED~\cite{chen2022cped} and MDPE~\cite{cai2025acmmm} integrate emotional context for richer behavioral cues. 
However, most existing datasets still lack comprehensive demographic coverage. Furthermore, as far as we know, the existing datasets with demographic annotations are based on the Big Five personality traits. 
To bridge this gap, we introduce a large-scale multimodal MBTI dataset DMSP, that explicitly annotates both traits and demographic attributes, enabling fairer multimodal personality understanding.

\begin{figure}[t]
  \centering
  \includegraphics[width=0.9\linewidth]{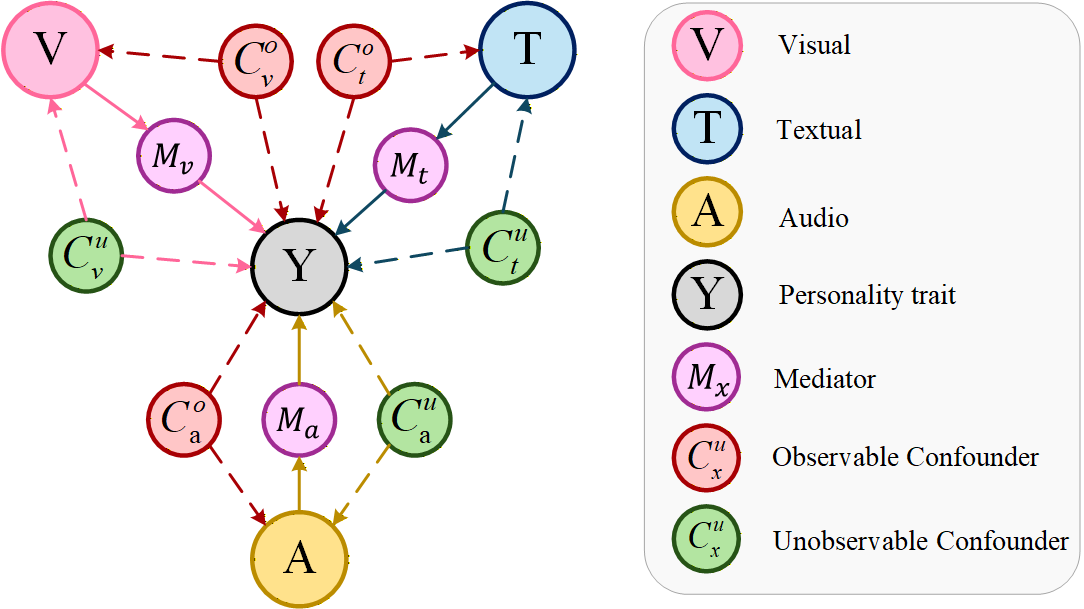}
  \caption{The proposed Structural Causal Model (SCM). The graph illustrates the causal relationships among multimodal inputs, modality mediators $M_x$, observable confounders $C_x^o$ (i.e., $C_v^o$, $C_a^o$, $C_t^o$), unobservable confounders $C_x^u$ (i.e., $C_v^u$, $C_a^u$, $C_t^u$), and the personality trait $Y$.}
  \Description{Structural Causal Model}
  \label{fig:SCM}
\end{figure}

\subsection{Causal Intervention}
Causal intervention modifies the distribution of independent variables to remove spurious or harmful dependencies on confounders, and is increasingly combined with deep learning to reduce spurious correlations across many domains \cite{deng2025deep}.

In computer vision, tasks such as image annotation  \cite{yang2021causal}  and face anti-spoofing \cite{long2023learning}  have been investigated using causal intervention to mitigate identity bias.
In multimodal learning, Liu et al. \cite{liu2024cimdd} designed back-door and front-door adjustments to model lexical, visual, and cross-modal confounders for fake news detection. 
CMCIR \cite{liu2024cmcir}  employed causal interventions to disentangle visual-linguistic spurious correlations for visual question answering.   
SuCI \cite{xu2024suci} is most relevant to our work, addressing subject variation in human multimodal understanding by causally disentangling subject-specific spurious correlations.

Inspired by the above research, we make the first attempt to design a customized causal intervention network to address subject bias in multimodal personality understanding.

\begin{figure*}[t]
  \centering
  \includegraphics[width=0.95\linewidth]{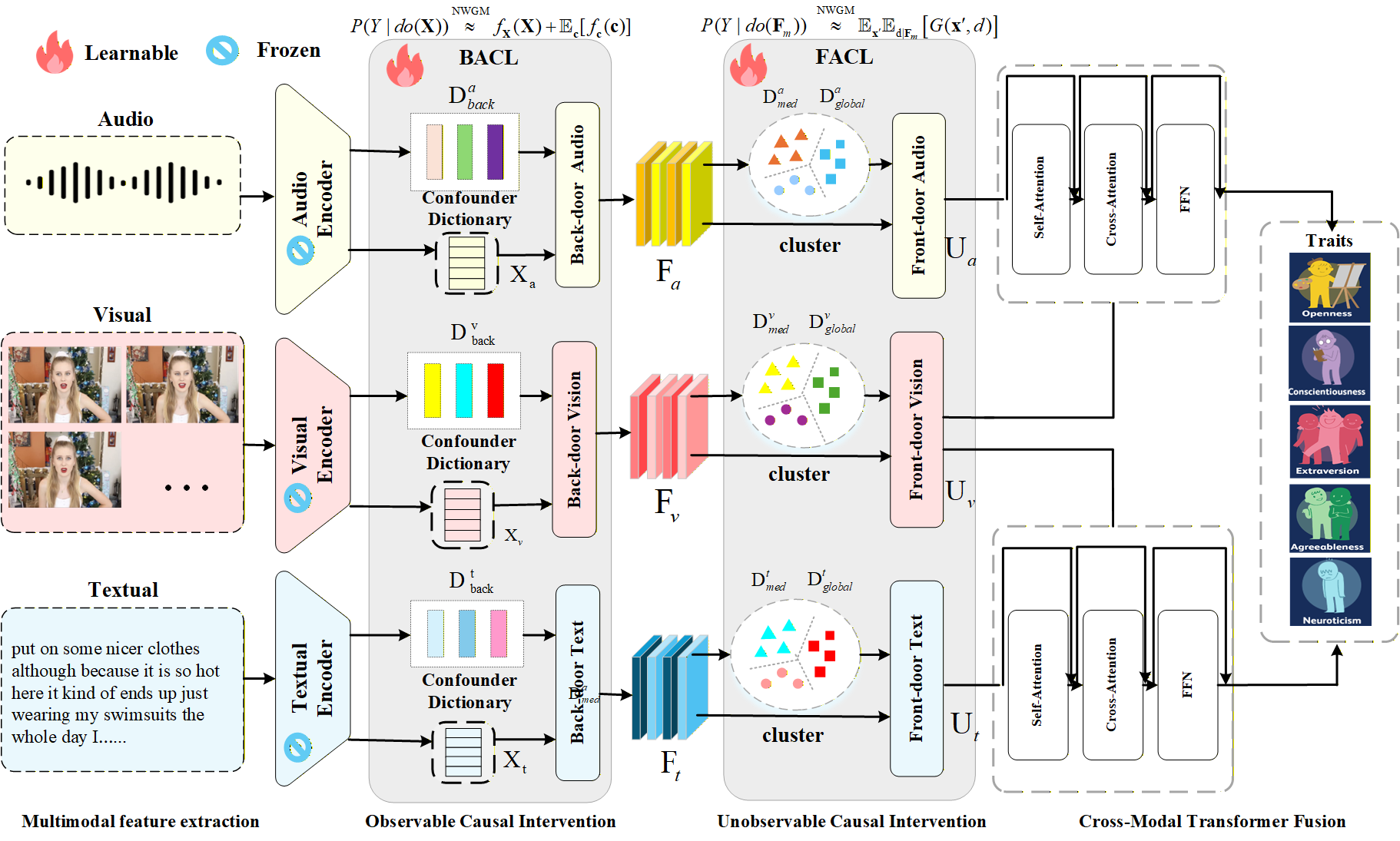}
  \caption{The architecture of our DCAN. Multimodal inputs $X$ are encoded and projected into a shared latent space $\mathbf{X}_m$. And then BACL eliminates observable confounders (e.g., demographics) via back-door adjustment  $\mathbf{F}_m$ with statistical prototype confounder dictionaries $\mathcal{D}_{back}^m$. FACL addresses unobservable confounders via front-door adjustment $\mathbf{U}_m$  by learning mediator $\mathcal{D}_{med}^m$ and global confounder dictionaries $\mathcal{D}_{global}^m$. Finally, the deconfounded multimodal representations $\mathbf{U}_m $ are fused by a cross-modal transformer for personality understanding.}
  \label{fig:model_overview}
\end{figure*}
\section{Task Definition}

The multimodal personality understanding task is formulated as a video-based multi-trait regression problem.Mathematically, given a user-generated video $X = \{V, A, T\}$ containing three 
modalities: the visual stream $V \in \mathbf{R}^{N_v \times d_v}$, 
the audio stream $A \in \mathbf{R}^{N_a \times d_a}$, and the textual stream 
$T \in \mathbf{R}^{N_t \times d_t}$, our goal is to learn a mapping $P(Y|X)$ to predict a target vector $Y = \{y_k\}_{k=1}^{K}$  of trait intensities score. Specifically, the output dimension is set to $K=5$ and $K=4$  for Big Five and MBTI soft-label regression, thereby unifying both tasks under a continuous prediction framework.

\section{Causal View at Multimodal Personality Understanding}

To investigate the causal mechanisms in multimodal personality understanding, we propose a Structural Causal Model (SCM) with variables: multimodal input $X$ (comprising visual, audio, and textual modalities), personality trait $Y$, modality-specific mediator $M_x$, observable confounder $C_x^o$, and unobservable confounder $C_x^u$, as shown in Figure~\ref{fig:SCM}. The causal graph is used to portray the study variables and their interactions through causal links. The link $X \to M_x$ reflects that the mediator extracts semantic representations from raw inputs, and the link $M_x \to Y$ reflects that the classifier predicts personality based on these representations. 

In particular, ubiquitous confounders affect both features and predictions. Observable confounders (e.g., demographics) introduce biases via $X \gets C_x^o \to Y$, while unobservable confounders (e.g., transient emotional states) affect the mediator via $X \gets C_x^u \to M_x$. We can intuitively identify the effect of $X$ on $Y$ in the graph, including the causal association ($X \to M_x \to Y$) and the confounding association. The confounding association can be a harmful shortcut, as models may learn spurious correlations caused by subject heterogeneity rather than intrinsic behavioral cues. In the causal association, the mediator $M_x$ is essential; we apply back-door adjustment to block $C_x^o$ and front-door adjustment via $M_x$ to handle $C_x^u$, thereby isolating the genuine causal effect.

\section{Dual Causal Adjustment Network}
As illustrated in Figure~\ref{fig:model_overview}, DCAN aims to disentangle spurious correlations from both observable and latent confounders, and achieve unbiased multimodal personality understanding. DCAN consists of four components:
(1) \textbf{Unimodal Feature Encoder} captures modality-specific representations;
(2) \textbf{Observable Demographic Causal Intervention} removes observable demographic confounders by explicitly confounder dictionary;
(3) \textbf{Unobservable Subject Confounding Causal Intervention} mitigates latent subject confounders by learning mediator-based intervention;
(4) \textbf{Cross-Modal Transformer Fusion} integrates the deconfounded multimodal representations for final prediction.

\subsection{Multimodal Feature Encoder}

Pre-trained encoders are leveraged to extract representations from heterogeneous data. Specifically, we utilize CLIP to generate frame-level visual embeddings $\mathbf{I}_v$ and transcript-based textual embeddings $\mathbf{I}_t$, while employing Wav2CLIP to extract audio features $\mathbf{I}_a$ from spectrograms. Formally,
\begin{equation}
\mathbf{I}_v = \Phi_{vis}(\mathbf{V}), \quad \mathbf{I}_a = \Phi_{aud}(\mathbf{A}), \quad \mathbf{I}_t = \Phi_{txt}(\mathbf{T}),
\label{eq:feature_extraction}
\end{equation}
where $\Phi_{vis}$ and $\Phi_{txt}$ denote the visual and textual encoders of CLIP, and $\Phi_{aud}$ denotes the Wav2CLIP audio encoder, respectively. We apply a linear projection to map these heterogeneous inputs ($\mathbf{I}_v, \mathbf{I}_t \in \mathbb{R}^{768}$, $\mathbf{I}_a \in \mathbb{R}^{512}$) into a unified $d$-dimensional latent space:
\begin{equation}
\mathbf{X}_m = \mathbf{I}_m \mathbf{W}_m^{p}, \quad m \in \{v, t, a\},
\label{eq:projection}
\end{equation}
where $\mathbf{W}_m^{p} \in \mathbb{R}^{d_m \times d}$ is a learnable projection matrix mapping raw features $\mathbf{I}_m$ into latent embeddings $\mathbf{X}_m \in \mathbb{R}^{d}$, which serve as the input for subsequent causal interventions.

\subsection{Observable Demographic Causal Intervention}
\label{sec:bacl}
\begin{figure}[h]
  \centering

  \begin{subfigure}[b]{1\linewidth}
    \centering
   
    \includegraphics[width=\linewidth]{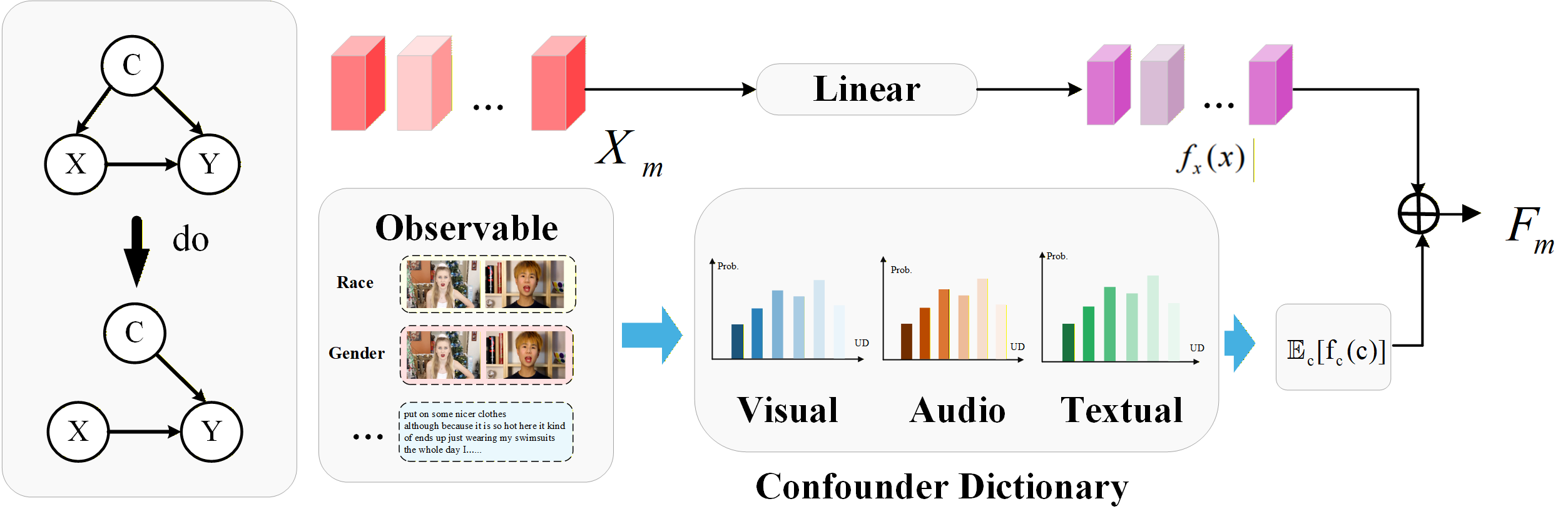}
 \caption{Back-door Adjustment Causal Learning (BACL)}
    \label{fig:bacl} 
  \end{subfigure}
  \begin{subfigure}[b]{1\linewidth}
    \centering
   
    \includegraphics[width=\linewidth]{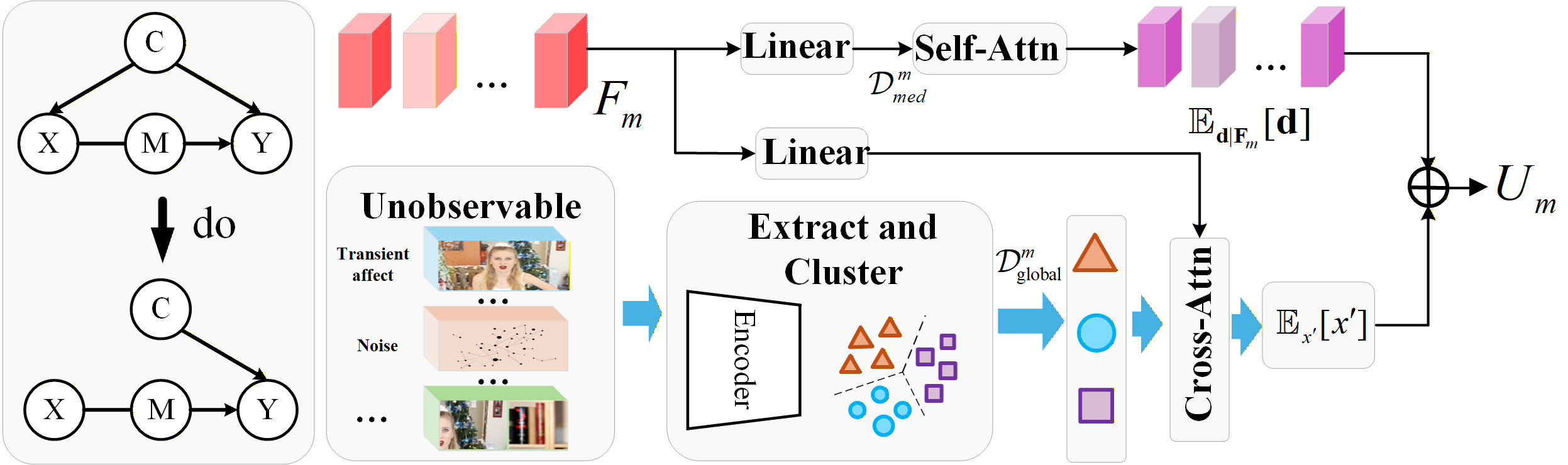}
    \caption{Front-door Adjustment Causal Learning (FACL)}
    \label{fig:facl} 
  \end{subfigure}
\caption{Illustration of the proposed causal intervention pipeline: (a) BACL targets observable confounders (e.g., demographics); (b) FACL targets unobservable confounders (e.g., subject emotions). }
  \label{fig:causal_framework}
\end{figure}
\textbf{Theoretical Formulation.}
The observational likelihood is typically formulated as $P(Y | \mathbf{X})=\sum_{\mathbf{c}} P(Y | \mathbf{X}, \mathbf{c}) P(\mathbf{c} | \mathbf{X})$, where $P(\mathbf{c} | \mathbf{X})$ introduces bias due to spurious correlations. To mitigate this, we employ the $do$-operator to sever the back-door path $\mathbf{c} \to \mathbf{X}$, deriving the causal effect as:
\begin{equation}
 P(Y | do(\mathbf{X})) = \sum_{\mathbf{c}} P(Y | \mathbf{X}, \mathbf{c}) P(\mathbf{c}).
 \label{eq:back-door_adjustment}
\end{equation}

\noindent\textbf{BACL Implementation.}
Subject demographics attributes (e.g., age, gender, race) are prominent observable confounders $\mathbf{c}$ in personality analysis. We first construct confounder dictionaries for each modality and then retrieval confounder to perform back-door adjustment for the intervention.

\noindent\textbf{Textual Modality.}
Based on the core principle of quantifying association bias, we construct a textual confounder dictionary $\mathcal{D}_{back}^t = \{\mathbf{c}_1^t, \ldots, \mathbf{c}_{K}^t\}$. Specifically, to identify words exhibiting significant probability differences across demographic groups (e.g., Gender, Age, Race), we compute the conditional probability $P(w|l)$ for each word $w$ under sensitive attribute $l$:
\begin{equation}
P(w|l)=\frac{N(w,l)}{N(l)+\epsilon}, 
\label{eq:word_prob}
\end{equation}
where $N(w,l)$ denotes the frequency of word $w$ in texts labeled $l$, and $N(l)$ is the total text count for label $l$. Words with scores exceeding a threshold are filtered to form $\mathcal{D}_{back}^t$~\cite{sap2014developing}.
 We then measure the distribution discrepancy using a bias score:
\begin{equation}
\text{bias\_score}(w)=\max_{l \in L}P(w|l) - \min_{l \in L}P(w|l).
\label{eq:bias_score}
\end{equation}

\noindent\textbf{Visual Modality.}
For the visual modality, we treat demographic attributes as observable confounders. We construct a dictionary $\mathcal{D}_{back}^v = \{\mathbf{c}_1^v, \ldots, \mathbf{c}_{K}^v\}$, where each prototype $\mathbf{c}_k^v \in \mathbb{R}^{d}$ corresponds to a unique triplet of (race, gender, age). CLIP~\cite{radford2021learning} is used to extract frame-level embeddings. During training, embeddings belonging to the same demographic combination are averaged to form a batch prototype $\tilde{\mathbf{c}}_k^v$. The global prototype is updated online via Exponential Moving Average (EMA)~\cite{he2020momentum}:
\begin{equation}
\mathbf{c}_k^{v(t)} \leftarrow \beta \mathbf{c}_k^{v(t-1)} + (1 - \beta)\tilde{\mathbf{c}}_k^{v(t)},
\label{eq:ema_visual}
\end{equation}
where $\beta$ is the momentum coefficient which ensures stable adaptation to the population distribution.

\noindent\textbf{Audio Modality.}
Unlike visual or textual data, audio contains transient semantic variations. We construct $\mathcal{D}_{back}^a = \{\mathbf{c}_1^a, \ldots, \mathbf{c}_{K}^a\}$ using the same demographic mapping as the visual modality. For an input sequence $\mathbf{X}_{seq} \in \mathbb{R}^{L \times d}$, we employ a cross-modal feature pooling mechanism to extract a global acoustic baseline (e.g., timbre, noise):
\begin{equation}
\mathbf{h}_{pool} = \sum_{t=1}^{L} \alpha_t \mathbf{x}_t,
\label{eq:audio_pooling}
\end{equation}
where $\alpha_t$ are learnable attention weights. The corresponding prototype $\mathbf{c}_{idx}^a$ (where $idx$ is the demographic index) is then updated via EMA similar to the visual modality:
\begin{equation}
\mathbf{c}_{idx}^{a(t)} \leftarrow \beta \mathbf{c}_{idx}^{a(t-1)} + (1 - \beta)\mathbf{h}_{pool}^{(t)}.
\label{eq:ema_audio}
\end{equation}

\noindent\textbf{Confounder Retrieval and Intervention.}
For a query feature $\mathbf{X}_m$, the modality-specific confounder representation $\mathbf{R}_{back}^m$ is retrieved from the corresponding dictionary $\mathcal{D}_{back}^m$ via attention:
\begin{equation}
\alpha_i = \mathrm{softmax}\left(\frac{(\mathbf{W}_Q^{b} \mathbf{X}_m)^\top (\mathbf{W}_K^{b} \mathbf{c}_i^m)}{\sqrt{d}}\right), \quad
\mathbf{R}_{back}^m = \sum_{i=1}^{K} \alpha_i \mathbf{c}_i^m.
\label{eq:confounder_retrieval}
\end{equation}

We approximate  intervention prediction $P(Y|do(\mathbf{X}))$ by the Normalized Weighted Geometric Mean (NWGM)\cite{zheng2022graph}:
\begin{equation} 
\begin{aligned} 
P(Y|do(\mathbf{X})) &= \sum_{\mathbf{c}} P(Y | \mathbf{X}, \mathbf{c}) P(\mathbf{c}) \\ 
&\stackrel{\text{NWGM}}{\approx} \mathbb{E}_{\mathbf{c}}\left[G(\mathbf{X}, \mathbf{c}) \right], 
\end{aligned} 
\label{eq:nwgm_backdoor} 
\end{equation}
where $G(\mathbf{X}, \mathbf{c}) = f_{\mathbf{X}}(\mathbf{X}) + f_{\mathbf{c}}(\mathbf{c})$ is the feature fusion function.  Consequently, the deconfounded representation can be simplified as:
 \begin{equation}
  P(Y|do(\mathbf{X})) \approx f_{\mathbf{X}}(\mathbf{X}) + \mathbb{E}_{\mathbf{c}}[f_{\mathbf{c}}(\mathbf{c})],
 \label{eq:bacl_approx}
\end{equation}
where $\mathbb{E}_{\mathbf{c}}[f_{\mathbf{c}}(\mathbf{c})]$ represents the expectation of the confounder context.   Specifically, we incorporate the causal context via a residual connection:
\begin{equation}
\mathbf{F}_m = \mathbf{X}_m + \gamma \cdot \mathbf{R}_{back}^m,
\label{eq:bacl_residual}
\end{equation}
where $\gamma$ is a learnable scaling parameter that dynamically regulates the injection of confounder information, yielding the bias-reduced representation $\mathbf{F}_m$.
\subsection{Unobservable Subject Confounding Causal Intervention}
\label{sec:facl}

\noindent\textbf{Theoretical Formulation.}
By inserting a mediator $\mathbf{M}$ to form the path $\mathbf{F}_m \to \mathbf{M} \to Y$, we apply the front-door adjustment rule to derive the causal effect:
\begin{equation}
P(Y|do(\mathbf{F}_m)) = \sum_{\mathbf{d}} P(\mathbf{d}|do(\mathbf{F}_m)) \sum_{\mathbf{x}'} P(Y|\mathbf{d}, \mathbf{F}_m=\mathbf{x}') P(\mathbf{x}'),
\label{eq:frontdoor_adjustment}
\end{equation}
where $\mathbf{x}'$ denotes potential input samples. We approximate this intervention in the feature space as $\mathbb{E}_{\mathbf{d} | \mathbf{F}_m}[\mathbf{d}] + \mathbb{E}_{\mathbf{x}'}[\mathbf{x}']$, corresponding to the $\oplus$ operator in Figure~\ref{fig:causal_framework}(b).

\noindent\textbf{FACL Implementation.}
To address unobservable confounders missed by BACL, we propose the Front-door Adjustment Causal Learning (FACL) module. The input to FACL is the deconfounded representation $\mathbf{F}_m$ of BACL.

\noindent\textbf{Mediator and Global Dictionaries.} To implement the front-door adjustment, we construct two dictionaries for each modality. Let $\mathcal{D}_{global}^m = \{\mathbf{g}_1^m, \ldots, \mathbf{g}_{N}^m\}$ denote the global feature dictionary approximating the input space $P(\mathbf{x}')$, and $\mathcal{D}_{med}^m = \{\mathbf{d}_1^m, \ldots, \mathbf{d}_K^m\}$ denote the mediator dictionary serving as the global knowledge prior. Specifically, both dictionaries are constructed by applying K-means clustering to the features of all training samples and sampling representative prototypes from each cluster, where $\mathbf{g}_k^m, \mathbf{d}_j^m \in \mathbb{R}^{d}$.

\noindent\textbf{NWGM Approximation.}
The goal is to compute the deconfounded representation $P(Y|do(\mathbf{F}_m))$. We can similarly approximate this by NWGM:
\begin{equation} 
\begin{aligned} 
P(Y|do(\mathbf{F}_m)) & = \sum_{\mathbf{d}} P(\mathbf{d} | \mathbf{F}_m) \sum_{\mathbf{x}'} P(Y|\mathbf{x}', \mathbf{d}) P(\mathbf{x}') \\ 
& \stackrel{\text{NWGM}}{\approx} \mathbb{E}_{\mathbf{x}'} \mathbb{E}_{\mathbf{d} | \mathbf{F}_m}\left[G'(\mathbf{x}', \mathbf{d}) \right], 
\end{aligned} 
\label{eq:nwgm_approx} 
\end{equation}
where $G'(\mathbf{x}', \mathbf{d})=\mathbf{W}_{g} \mathbf{x}' + \mathbf{W}_{d} \mathbf{d}$ is a linear layer for feature fusion. $\mathbf{W}_{g}, \mathbf{W}_{d} \in \mathbb{R}^{d \times d}$ are learnable weight matrices. Consequently, the deconfounded representation $\mathbf{U}_m$ can be simplified into two expectation terms:
\begin{equation}
\mathbf{U}_m = \mathbf{W}_{g} \cdot \mathbb{E}_{\mathbf{x}'}[\mathbf{x}'] + \mathbf{W}_{d} \cdot \mathbb{E}_{\mathbf{d} | \mathbf{F}_m}[\mathbf{d}].
\label{eq:facl_output}
\end{equation}

\noindent\textbf{Dual-Attention Adjustment.}
Directly estimating the prior distributions for $\mathbb{E}_{\mathbf{x}'}[\mathbf{x}']$ and $\mathbb{E}_{\mathbf{d} | \mathbf{F}_m}[\mathbf{d}]$ is infeasible due to the high dimensionality of the sample space. Therefore, we adopt a query-based attention mechanism to approximate these expectations. Specifically, we generate two query vectors $\mathbf{Q}_{g} = \mathbf{F}_m \mathbf{W}_Q^{g}$ and $\mathbf{Q}_{d} = \mathbf{F}_m \mathbf{W}_Q^{d}$ from the BACL output $\mathbf{F}_m$. The expectations are then computed by retrieving relevant prototypes from the dictionaries:
\begin{equation}
\mathbb{E}_{\mathbf{d} | \mathbf{F}_m}[\mathbf{d}] = \sum_{j=1}^{K} \mathrm{softmax}\left(\frac{\mathbf{Q}_{d} (\mathbf{K}_{d})^\top}{\sqrt{d}}\right)_{j} \cdot \mathbf{d}_j^m,
\label{eq:mediator_expectation}
\end{equation}
\begin{equation}
\mathbb{E}_{\mathbf{x}'}[\mathbf{x}'] = \sum_{k=1}^{N} \mathrm{softmax}\left(\frac{\mathbf{Q}_{g} (\mathbf{K}_{g})^\top}{\sqrt{d}}\right)_{k} \cdot \mathbf{g}_k^m,
\label{eq:global_expectation}
\end{equation}
where $\mathbf{K}_d = \mathbf{W}_K^d [\mathbf{d}_1^m; \ldots; \mathbf{d}_K^m]$ and $\mathbf{K}_{g} = \mathbf{W}_K^g [\mathbf{g}_1^m; \ldots; \mathbf{g}_N^m]$ are key projections of the mediator dictionary $\mathcal{D}_{med}^m$ and global dictionary $\mathcal{D}_{global}^m$, respectively.
By substituting these estimates back into Eq.~\eqref{eq:facl_output}, we obtain the final causality-enhanced representation $\mathbf{U}_m$ through complete front-door adjustment.
\subsection{Cross-Modal Transformer Fusion}
\label{sec:fusion}

To integrate deconfounded multimodal representations
$\{\mathbf{U}_v, \mathbf{U}_a, \mathbf{U}_t\}$, we first compute pairwise cross-modal similarities using cosine similarity to capture modality-consistent semantics.
The visual–text similarity $\mathbf{S}_{vt}$ and visual–audio similarity $\mathbf{S}_{va}$ are concatenated with the visual features to form the input for multi-head attention fusion:
$\mathbf{O}_{fused} = [\mathbf{U}_v; \mathbf{S}_{vt}; \mathbf{S}_{va}] \in \mathbb{R}^{B \times 3 \times d}$.
The final personality traits are obtained through global average pooling and an MLP head:
\begin{equation}
\hat{\mathbf{Y}} = \sigma\!\left(\mathrm{MLP}\!\left(\mathrm{MeanPooling}(\mathbf{O}_{fused})\right)\right),
\label{eq:prediction}
\end{equation}
where $\hat{\mathbf{Y}}$ denotes the predicted personality trait scores, and $\sigma(\cdot)$ represents the task-specific activation function.
For both Big Five and MBTI taxonomy, we apply a sigmoid activation to bound the outputs within $[0,1]$,
as the MBTI task adopts \textit{soft labels} that reflect continuous type probabilities rather than hard categorical assignments.

\noindent\textbf{Training Objective.}
The model is optimized by a unified loss function combining Mean Squared Error (MSE) loss and $L_2$ regularization:
\begin{equation}
\mathcal{L} = \frac{1}{N}\sum_{i=1}^{N}\sum_{k=1}^{K}(y_{i,k} - \hat{y}_{i,k})^2 + \lambda \sum_{\theta \in \Theta} \|\theta\|_2^2,
\label{eq:total_loss}
\end{equation}
where $N$ is the batch size, $K$ is the number of personality dimensions ($K=5$ for Big Five, $K=4$ for MBTI), $y_{i,k}$ and $\hat{y}_{i,k}$ denote the ground-truth and predicted scores for the $k$-th trait of the $i$-th sample, and $\lambda$ is the regularization coefficient.

\section{Experiments}
\begin{table*}[t] 
 \centering 
 \caption{Comprehensive Performance Comparison of Models on CFI-V2 and DMSP datasets. Best results are \textbf{bolded}.} 
 \label{tab:all_performance} 
 \resizebox{\textwidth}{!}{ 
 \setlength{\tabcolsep}{4pt} 
 \renewcommand{\arraystretch}{1.15} 
 \begin{tabular}{lcccccccccccccc} 
 \toprule 
 \multirow{2}{*}{\textbf{Model}} & \multicolumn{9}{c}{\textbf{CFI-V2 (Big Five)}} & \multicolumn{5}{c}{\textbf{DMSP (MBTI)}} \\ 
 \cmidrule(lr){2-10} \cmidrule(lr){11-15} 
 & \textbf{ACC} & \textbf{PCC} & \textbf{CCC} & \textbf{$R^2$} & \textbf{Ext} & \textbf{Agr} & \textbf{Con} & \textbf{Neu} & \textbf{Ope} & \textbf{E/I} & \textbf{N/S} & \textbf{F/T} & \textbf{J/P} & \textbf{ACC} \\ 
 \midrule 
 CNN-LSTM~\cite{subramaniam2016bi} & 0.9151 & 0.6874 & 0.6517 & 0.4736 & 0.9148 & 0.9131 & 0.9207 & 0.9121 & 0.9146 & 0.8861 & 0.9047 & 0.8848 & 0.9283 & 0.9010 \\ 
 PersEmoN~\cite{zhang2019persemon} & 0.9160 & 0.6969 & 0.6739 & 0.4790 & 0.9158 & 0.9161 & 0.9184 & 0.9151 & 0.9145 & 0.9045 & 0.8605 & 0.7640 & 0.9421 & 0.8678 \\
DAN~\cite{wei2017deep}& 0.9174 & 0.7031 & 0.6703 & 0.4962 & 0.9190 & 0.9147 & 0.9221 & 0.9160 & 0.9152 & 0.8945 & 0.9149 & 0.8851 & 0.9266 & 0.9053 \\ 
  \hline 
 ResNet-Feature~\cite{liao2024benchmark} & 0.9150 & 0.6877 & 0.6622 & 0.4701 & 0.9175 & 0.9118 & 0.9214 & 0.9115 & 0.9128 & 0.8867 & 0.8404 & 0.7984 & 0.9233 & 0.8622 \\ 
 ResNet-Score~\cite{liao2024benchmark} & 0.9120 & 0.6690 & 0.6475 & 0.4296 & 0.9130 & 0.9107 & 0.9179 & 0.9101 & 0.9084 & 0.7926 & 0.9139 & 0.8810 & 0.6840 & 0.8179 \\ 
  ResNet~\cite{guccluturk2016deep} & 0.9172 & 0.7082 & 0.6729 & 0.4992 & 0.9185 & 0.9162 & 0.9219 & 0.9138 & 0.9157 & 0.8877 & 0.9076 & 0.8849 & 0.9258 & 0.9015 \\ 
 \hline 
 CRNet~\cite{li2020crnet} 
& 0.9188 & 0.7170 & 0.6865 & 0.5147 & 0.9193 & 0.9173 & 0.9226 & 0.9173 & 0.9175 & 0.9132 & 0.8587 & 0.7710 & 0.9388 & 0.8754 \\
 MM-ResVGG \cite{suman2022multi}
 & 0.9197 & 0.7269 & 0.6881  & 0.5267 & 0.9177 & 0.9247 & 0.9211 & 0.9173 & 0.9176 & 0.9448 & 0.9190 & 0.8692 & 0.9346 & 0.9169 \\ 
PMGRT~\cite{wang2025novel}  &0.9199 & 0.7304 & 0.7130 & 0.5286 & 0.9191 & 0.9218 & 0.9226 & 0.9185 & 0.9176 & 0.8816 & 0.9008 & 0.8847 & 0.9261 & 0.8983\\
 \midrule 
 \textbf{DCAN (Ours)} & \textbf{0.9211} & \textbf{0.7377} & \textbf{0.7137} & \textbf{0.5446} & \textbf{0.9194} & \textbf{0.9267} & \textbf{0.9227} & \textbf{0.9187} & \textbf{0.9181} & \textbf{0.9572} & \textbf{0.9322} & \textbf{0.8860} & \textbf{0.9408} & \textbf{0.9290} \\ 
 \bottomrule 
 \end{tabular} 
 } 
 \end{table*}
 
\subsection{Datasets} 
\label{sec:dataset}
 \noindent\textbf{DMSP (Ours)}: We construct a Demographic-annotated Multimodal Student Personality (DMSP) dataset to support fairness multimodal personality understanding. It comprises 839 dyadic interaction videos derived from adolescent student participants. DMSP distinguishes itself through three key design features. First, it ensures high-quality labels by employing a rigorous psychometric assessment protocol (based on a simplified 28-item questionnaire), avoiding the self-reporting noise typical of social media datasets. Second, we provide controllable demographic annotations by explicitly labeling sensitive attributes such as gender and age, whose proportions can be flexibly adjusted to form either balanced or intentionally imbalanced demographic distributions, enabling controlled fairness evaluation under varying demographic settings. Finally, the dataset adopts trait-based regression via soft labels, which allows for fine-grained behavioral analysis of intermediate personality states.

\noindent\textbf{CFI-V2}: This widely-used benchmark focuses on apparent personality traits from short video blogs, providing aligned visual, audio, and textual data.
However, it exhibits a long-tail distribution in demographic attributes and lacks explicit fairness-aware annotations, rendering it insufficient for rigorous bias mitigation studies.

\subsection{Baselines} 
\label{sec:dataset}

We compare our proposed DCAN with the following three categories of representative models.
\noindent\textbf{(1) DNN Models:} 
\textbf{CNN-LSTM} \cite{Subramaniam2016} employs segment-based temporal modeling networks to capture coarse-grained personality cues from audio-visual streams. 
\textbf{DAN} \cite{Wei2018} utilizes bimodal regression to map audiovisual cues to personality traits. 
\textbf{PersEmoN} \cite{zhang2019persemon} jointly models personality and emotion via shared feature representation to exploit task correlations.
\noindent\textbf{(2) ResNet Family:} Our evaluation includes two fusion variants of \textbf{ResNet} \cite{guccluturk2016deep}. 
\textbf{ResNet-Feature} \cite{liao2024open} employs dual ResNet1D branches to extract modality features, followed by feature-level concatenation for early fusion. 
\textbf{ResNet-Score} \cite{liao2024open} performs late-stage fusion via decision-level score averaging.
\noindent\textbf{(3) Trimodal Fusion Models:} We also compare against state-of-the-art architectures that aggregate video, audio, and text features. 
\textbf{CRNet} \cite{Li2020} designs a three-stream classification-regression network to capture latent features across modalities. 
\textbf{MM-ResVGG} \cite{suman2022multi} integrates ResNet, VGGish, and n-gram CNN for trimodal feature fusion. 
\textbf{PMGRT} \cite{wang2025novel} introduces a graph relational transformer to capture long-term temporal dependencies and cross-modal associations.

\subsection{Implementation Details}
\label{sec:implementation_details}

All models follow consistent feature extraction procedures to ensure fair comparison. Visual and textual features are extracted using frozen CLIP encoders, yielding 768-dimensional embeddings. For the acoustic modality, we employ Wav2CLIP to obtain 512-dimensional audio representations.For the causal modules, the dictionary sizes for the confounder (BACL) and mediator (FACL) are set to \(K = 128\) and \(M = 64\), respectively. The framework is optimized with AdamW using a batch size of 32 and an initial learning rate of \(1 \times 10^{-4}\), following a cosine annealing schedule. All experiments are conducted on an NVIDIA RTX 4090D GPU using PyTorch 2.3.1. Reported results are the mean of three independent runs. 
\subsection{Metrics}
\label{subsec:metrics}

Following prior work~\cite{liao2024open}, we evaluate model performance using \textit{Accuracy (ACC)} defined as the inverse mean absolute error, \textit{Pearson and Concordance Correlation Coefficients (PCC/CCC)} to measure correlation and agreement, and the \textit{Coefficient of Determination ($R^2$)} to quantify explained variance.

To evaluate the fairness of personality understanding across different demographic groups\cite{islam2024fairness}, we adopt \textit{Demographic Parity (DP)} to measure the statistical independence between the prediction trait and sensitive attributes.
\begin{equation}
\mathrm{DP} = \left| \mathbb{E}[\hat{Y} \mid A = a] - \mathbb{E}[\hat{Y} \mid A = b] \right|,
\label{eq:a}
\end{equation}
where $A$ denotes a sensitive attribute (e.g., gender or age group), and $\hat{Y}$ represents the predicted personality score.  \textit{Equal Opportunity (EO)} measures the consistency of true positive trait rates across different demographic groups.
\begin{equation}
\mathrm{EO} = \left| \mathbb{E}[\hat{Y} \mid Y \ge \tau, A = a] - \mathbb{E}[\hat{Y} \mid Y \ge \tau, A = b] \right|,
\label{eq:b}
\end{equation}
where threshold $\tau$ on the ground-truth label $Y$ to identify positive samples ($Y \ge \tau$).

\subsection{Experimental Results}
\label{sec:exp_results}

\noindent\textbf{(1) Overall Performance.} As shown in Table~\ref{tab:all_performance}, DCAN achieves 0.9211 ACC, 0.7377 PCC, 0.7137 CCC, and 0.5446 $R^2$ on the CFI-V2 dataset and improves over the previous best method (PMGRT) by 0.12\% in ACC, 0.73\% in PCC, 0.07\% in CCC, and 1.60\% in $R^2$.
On the DMSP dataset, DCAN achieves an overall accuracy of 0.9290, representing a 1.21\% absolute improvement over the previous SOTA (MM-ResVGG). These consistent gains across both datasets highlight the effectiveness of the proposed causal deconfounding framework in enhancing personality recognition performance.

\noindent\textbf{(2) Analysis of Performance Gains.}
The performance gains of DCAN come from explicitly modeling and mitigating confounding factors that obscure personality cues.
Personality traits are often entangled with subject variables such as age, gender, and race, causing models to learn spurious correlations.
DCAN simultaneously addresses both observable and latent confounders through dual causal deconfounding. 
This end-to-end causal disentanglement enables the capture of causally relevant and subject-unbiased personality representations, thereby improving the reliability.

\noindent\textbf{(3)  Observations of Specific Dataset.}
The performance improvement is more pronounced on the DMSP dataset (1.21\%) than on CFI-V2 (0.12\%), indicating that DCAN is particularly effective in datasets with stronger demographic biases. DMSP, which exhibits greater heterogeneity in age and gender distributions, tends to induce shortcut learning in baseline models that rely on superficial correlations.
By performing fine-grained causal intervention and invariant feature optimization, DCAN effectively mitigates such confounding effects, allowing the model to learn personality-related representations that are less influenced by subject shortcuts. These findings substantiate our hypothesis that causal deconfounding mechanisms are critical for disentangling intrinsic personality signals from both observable and latent confounders in multimodal personality understanding.

\subsection{Ablation Studies}
The tables  \ref{tab:ablation_final} and \ref{tab:ablation_dmsp_final} show the evaluation of the effectiveness of removing specific modules and modes from the complete DCAN framework on the DMSP and CFI-V2 datasets.

\begin{table}[t] 
   \caption{Ablation study of key components on the CFV-V2 dataset. Best results are \textbf{bolded}.} 
   \label{tab:ablation_final} 
   \resizebox{\linewidth}{!}{ 
   \begin{tabular}{l cccc} 
     \toprule 
     \textbf{Model Variant} & \textbf{ACC} & \textbf{PCC} & \textbf{CCC} & \textbf{$R^2$} \\ 
     \midrule 
     \textbf{DCAN (Full)} & \textbf{0.9211} & \textbf{0.7377} & \textbf{0.7137} & \textbf{0.5446} \\ 
     \midrule 
     \multicolumn{5}{l}{\textit{Causal Modules Ablation}} \\ 
     w/o BACL & 0.9165 & 0.7166 & 0.6923 & 0.5351 \\ 
     w/o FACL & 0.9160 & 0.7100 & 0.6650 & 0.5236 \\ 
     w/o BACL \& FACL & 0.9147 & 0.7075 & 0.6588 & 0.5454 \\ 
     \midrule 
     \multicolumn{5}{l}{\textit{Modality Ablation}} \\ 
     Only Vision & 0.9135 & 0.6964 & 0.6769 & 0.5021 \\ 
     Only Text & 0.8792 & 0.2574 & 0.1974 & 0.0702 \\ 
     Only Audio & 0.8905 & 0.4463 & 0.3905 & 0.1682 \\ 
     Vision \& Text       & 0.9136 & 0.7007 & 0.6784 & 0.5164 \\ 
     Vision \& Audio       & 0.9156 & 0.7128 & 0.6932 & 0.5392 \\ 
     Text \& Audio      & 0.8913 & 0.4691 & 0.4153 & 0.1445 \\ 
   \bottomrule 
 \end{tabular} 
 } 
 \end{table} 
\begin{table}[t] 
   \caption{Ablation study of key components on the DMSP dataset.  Best results are \textbf{bolded}.} 
   \label{tab:ablation_dmsp_final} 
   \resizebox{\linewidth}{!}{ 
   \begin{tabular}{l ccccc} 
     \toprule 
     \textbf{Model Variant} & \textbf{ACC} & \textbf{E/I} & \textbf{N/S} & \textbf{F/T} & \textbf{J/P} \\ 
     \midrule 
     \textbf{DCAN (Full)} & \textbf{0.9290} & \textbf{0.9572} & \textbf{0.9322} & \textbf{0.8860} & \textbf{0.9408} \\ 
     \midrule 
     \multicolumn{6}{l}{\textit{Causal Modules Ablation}} \\ 
     w/o BACL & 0.9266 & 0.9560 & 0.9299 & 0.8797 & 0.9406 \\ 
     w/o FACL & 0.9194 & 0.9570 & 0.9109 & 0.8588 & 0.9395 \\ 
     w/o BACL \& FACL & 0.9092 & 0.9517 & 0.8948 & 0.8397 & 0.9385 \\ 
     \midrule 
     \multicolumn{6}{l}{\textit{Modality Ablation}} \\ 
     Only Vision  & 0.9267 & 0.9562 & 0.9316 & 0.8811 & 0.9360 \\ 
     Only Text  & 0.9251 & 0.9571 & 0.9317 & 0.8734 & 0.9359 \\ 
     Only Audio & 0.9245 & 0.9565 & 0.9318 & 0.8724 & 0.9360 \\ 
     Vision \& Audio       & 0.9254 & 0.9569 & 0.9286 & 0.8752 & 0.9398 \\ 
     Vision \& Text     & 0.9248 & 0.9568 & 0.9302 & 0.8788 & 0.9329 \\ 
     Text \& Audio       & 0.9225 & 0.9559 & 0.9270 & 0.8691 & 0.9380 \\ 
   \bottomrule 
 \end{tabular} 
 } 
 \end{table}

\noindent\textbf{Effectiveness of Causal Modules.} 
Removing the FACL module causes the most significant degradation on CFI-V2, with ACC dropping from 0.9211 to 0.9160 and CCC from 0.7137 to 0.6650, indicating that latent unobserved confounders have a substantial effect on model performance.       Removing BACL also reduces performance, which confirms that observable demographic confounders distort personality modeling. When both modules are removed, the degradation is most pronounced (ACC: 0.9147, CCC: 0.6588). On DMSP, similar trends are observed—overall accuracy drops from 0.9290 to 0.9092, with the F/T dimension showing the largest decline (0.8860 to 0.8397), validating the necessity of causal adjustment for reducing demographic and contextual biases.

\textbf{Contribution of  Modalities.} 
On CFI-V2, vision proves to be the most informative single modality (ACC: 0.9135, PCC: 0.6964), while text alone performs the worst  (ACC: 0.8792, PCC: 0.2574). Among bimodal settings, Vision \& Audio achieves the best results (ACC: 0.9156, PCC: 0.7128). On DMSP, sdifferences between single modalities are smaller (around 0.925 ACC), yet the full trimodal DCAN consistently achieves the highest performance, highlighting that integrating visual, audio, and textual streams captures richer personality representation.

\subsection{Fairness Analysis}
\label{app:fairness_comparison}
As shown in Table \ref{tab:fairness_final},  we evaluate the de-biasing capability of DCAN with each demographic group, using the fairness metrics Demographic Parity (DP) and Equal Opportunity (EO), as defined in Equations \ref{eq:a} and \ref{eq:b}. Race fairness is evaluated only on CFI-V2 due to the lack of race annotations in DMSP.

\begin{table*}[h]
\centering
\caption{Comprehensive fairness comparison on CFI-V2 and DMSP datasets. $\downarrow$ indicates lower is better.}
\label{tab:fairness_final}
\resizebox{\textwidth}{!}{%
\begin{tabular}{l|cccccc|cccccccc}
\toprule
\multirow{3}{*}{\textbf{Model}} & \multicolumn{6}{c|}{\textbf{DMSP (MBTI)}} & \multicolumn{8}{c}{\textbf{CFI-V2 (Big Five)}} \\
 & \multicolumn{2}{c}{\textbf{Overall}} & \multicolumn{2}{c}{\textbf{Gender}} & \multicolumn{2}{c|}{\textbf{Age}} & \multicolumn{2}{c}{\textbf{Overall}} & \multicolumn{2}{c}{\textbf{Gender}} & \multicolumn{2}{c}{\textbf{Race}} & \multicolumn{2}{c}{\textbf{Age}} \\
 & \textbf{DP$\downarrow$ } & \textbf{EO$\downarrow$ } & \textbf{DP$\downarrow$ } & \textbf{EO$\downarrow$ } & \textbf{DP$\downarrow$ } & \textbf{EO$\downarrow$ } & \textbf{DP$\downarrow$ } & \textbf{EO$\downarrow$ } & \textbf{DP$\downarrow$ } & \textbf{EO$\downarrow$ } & \textbf{DP$\downarrow$ } & \textbf{EO$\downarrow$ } & \textbf{DP$\downarrow$ } & \textbf{EO$\downarrow$ } \\ \midrule
CNN-LSTM \cite{subramaniam2016bi} & 0.0653 & 0.0767 & 0.0286 & 0.0643 & 0.1673 & 0.1657 & 0.3328 & 0.2240 & 0.0616 & 0.0532 & 0.2286 & 0.1628 & 0.7082 & 0.4560 \\
DAN \cite{wei2017deep} & 0.0941 & 0.0908 & 0.0322 & 0.0521 & 0.2500 & 0.2201 & 0.3776 & 0.2221 & 0.0963 & 0.0788 & 0.3084 & 0.1628 & 0.7282 & 0.4246 \\
PersEmoN \cite{zhang2019persemon} & 0.0905 & 0.0997 & 0.0214 & 0.0490 & 0.2500 & 0.2500 & 0.3555 & 0.2047 & 0.0994 & 0.0795 & 0.2479 & 0.1010 & 0.7192 & 0.4337 \\ \midrule
ResNet \cite{guccluturk2016deep} & 0.0778 & 0.0962 & 0.0107 & 0.0387 & 0.2227 & 0.2500 & 0.3192 & 0.2770 & 0.1291 & 0.1151 & 0.2219 & 0.1316 & 0.6067 & 0.5844 \\
ResNet18 (Score) \cite{liao2024benchmark} & 0.0802 & 0.1026 & 0.0179 & 0.1005 & 0.2228 & 0.2072 & 0.2716 & 0.3321 & 0.0971 & 0.0850 & 0.2172 & 0.1370 & 0.5004 & 0.7744 \\
ResNet18 (Feat.) \cite{liao2024benchmark} & 0.0802 & 0.1026 & 0.0179 & 0.1005 & 0.2228 & 0.2072 & 0.3089 & 0.2949 & 0.1023 & 0.0903 & 0.2480 & 0.1121 & 0.5765 & 0.6823 \\ \midrule
CRNet \cite{li2020crnet} & 0.0656 & 0.0864 & 0.0214 & 0.0597 & 0.1754 & 0.1995 & 0.3277 & 0.2378 & 0.0840 & 0.0643 & 0.2410 & 0.1480 & 0.6580 & 0.5013 \\
MM-ResVGG \cite{suman2022multi} & 0.0749 & 0.0799 & 0.0250 & 0.0400 & 0.1997 & 0.1997 & 0.3096 & 0.3004 & 0.0914 & 0.0745 & 0.2519 & 0.1106 & 0.5854 & 0.7162 \\
PMGRT \cite{wang2025novel} & 0.0747 & 0.1064 & 0.0286 & 0.1236 & 0.1955 & 0.1955 & 0.2157 & 0.2025 & 0.0702 & 0.0735 & 0.1831 & 0.1491 & 0.3937 & 0.3849 \\ \midrule
\textbf{DCAN (Ours)} & \textbf{0.0522} & \textbf{0.0649} & \textbf{0.0086} & \textbf{0.0327} & \textbf{0.1336} & \textbf{0.1402} & \textbf{0.1985} & \textbf{0.1892} & \textbf{0.0584} & \textbf{0.0496} & \textbf{0.1750} & \textbf{0.0910} & \textbf{0.3521} & \textbf{0.3418} \\ \bottomrule
\end{tabular}%
}
\end{table*}

\noindent\textbf{(1) Fairness on Each Demographic Group.}
We separately evaluate the statistical independence and opportunity fairness of the model predictions with respect to sensitive demographic attributes, including gender, age, and race.
Regarding \textbf{gender}, DCAN effectively mitigates spurious correlations, achieving near-perfect fairness on DMSP (DP 0.0086) compared to ResNet (DP 0.0107), and superior results on CFI-V2 (DP 0.0584) compared to CNN-LSTM (DP 0.0616). 
Regarding \textbf{age}, which is particularly challenging in personality computing, DCAN demonstrates substantial improvement. It lowers Age DP to 0.3521 on CFI-V2, outperforming the strong baseline PMGRT (DP 0.3937). Similarly, on DMSP, DCAN achieves an Age DP of 0.1336 and EO of 0.1402, surpassing CNN-LSTM (DP 0.1673). These results prove that our causal intervention strategy effectively isolates both gender-specific and age-specific confounding effects.
Regarding \textbf{race}, DCAN maintains a superior balance with a DP of 0.1750 and EO of 0.0910, outperforming the competitive baseline PMGRT (DP 0.1831). These results confirm that the BACL module successfully adjusts for observable demographic shifts, ensuring equitable outcomes across racial groups.

\noindent\textbf{(2) Overall Fairness Performance.}
 DCAN achieves the lowest overall DP (0.1985) and EO (0.1892), outperforming the previous SOTA PMGRT (DP 0.2157)  and (EO 0.2025) on the CFI-V2 dataset.  Similarly, DCAN further demonstrates superior fairness with an overall DP of 0.0522 and EO of 0.0649, surpassing the best baseline CNN-LSTM (DP 0.0653) and (EO 0.0767) on the DMSP dataset, highlighting the importance of addressing both observable and latent confounders.

\subsection{ Out-of-Distribution Generalization Analysis}
\label{sec:demographic_evaluation}
\begin{figure*}[t]
  \centering
  \begin{subfigure}[b]{0.3\textwidth}
    \centering
    \includegraphics[width=\linewidth]{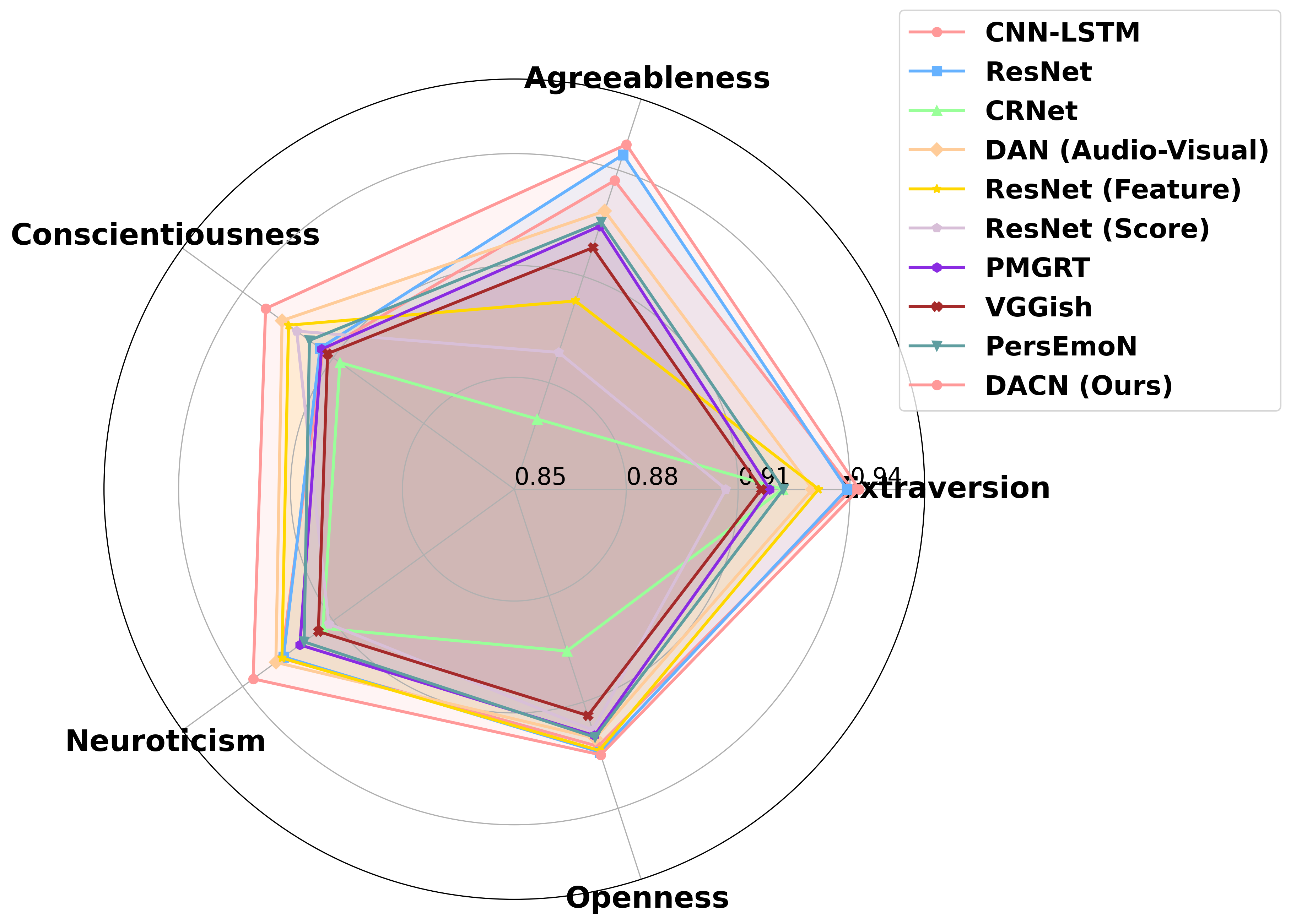}
    \caption{Age}
  \end{subfigure}
  \hfill
  \begin{subfigure}[b]{0.3\textwidth}
    \centering
    \includegraphics[width=\linewidth]{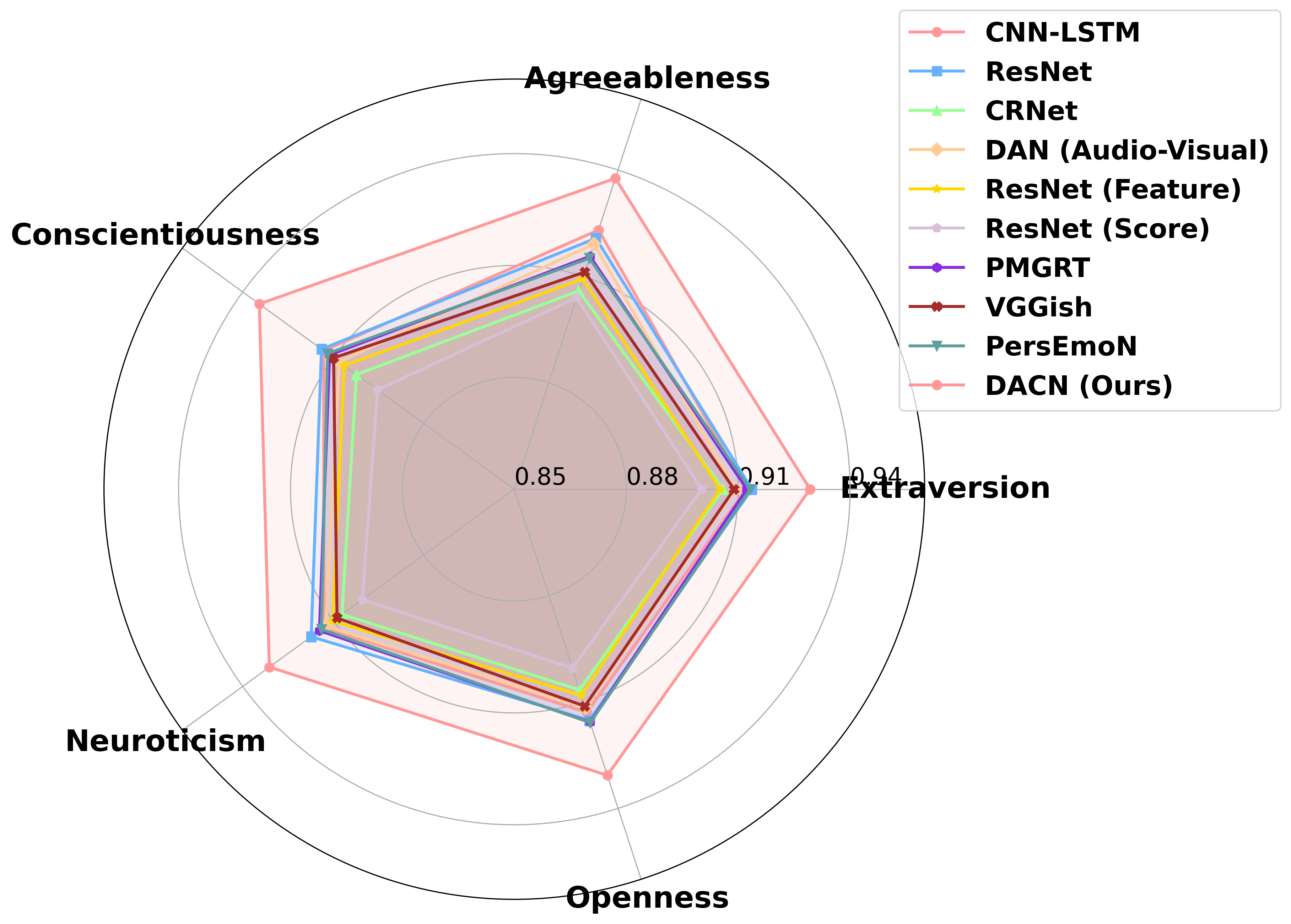}
    \caption{Gender}
  \end{subfigure}
  \hfill
  \begin{subfigure}[b]{0.3\textwidth}
    \centering
    \includegraphics[width=\linewidth]{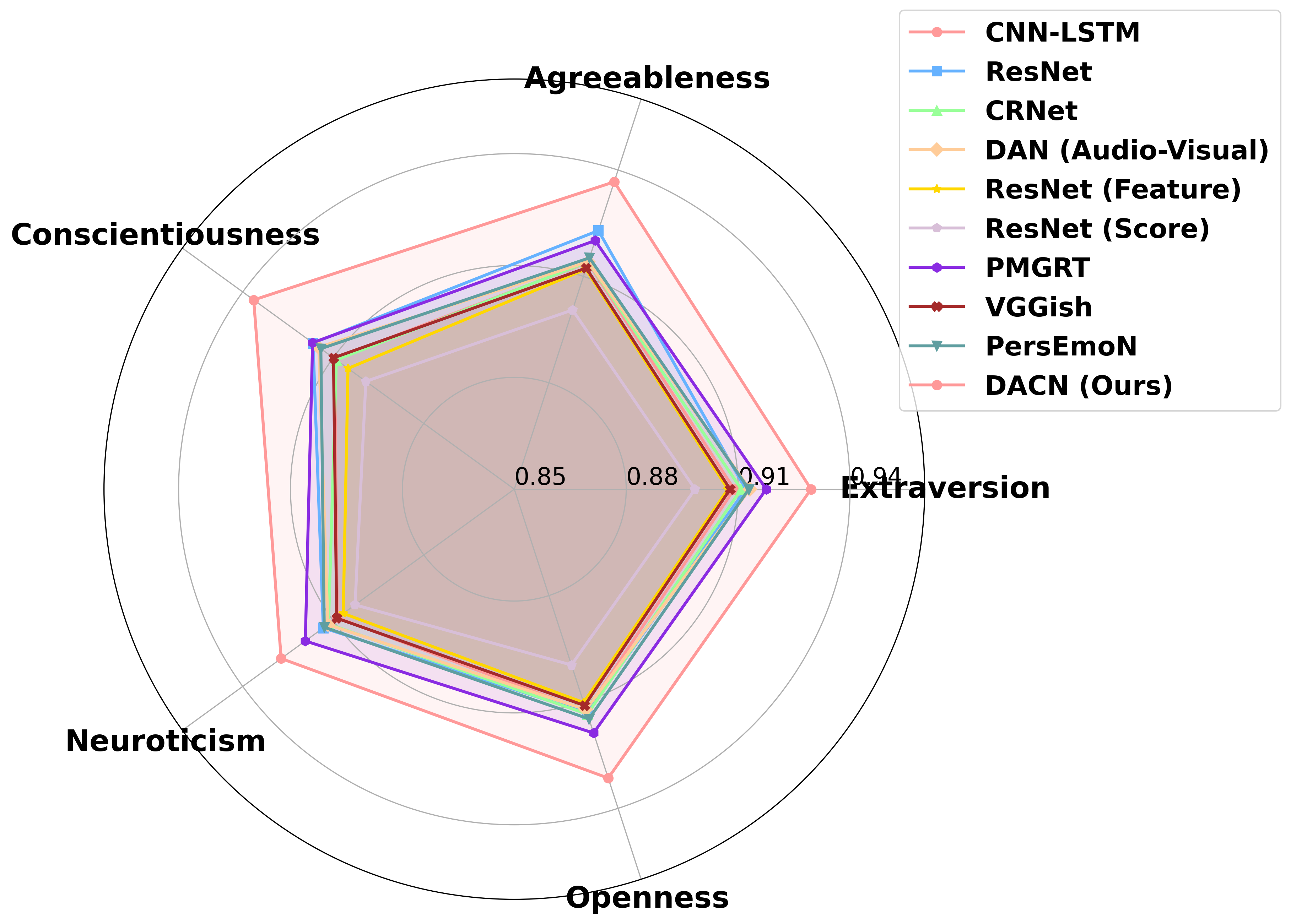}
    \caption{Ethnicity}
  \end{subfigure}
  
  \caption{Out-of-distribution (OOD) evaluation on CFI-V2 across demographic attributes.}
  \label{fig:demographic_radars_cfi}
\end{figure*}

\noindent\textbf{(1) Out-of-Distribution on Big Five Traits.} Figure~\ref{fig:demographic_radars_cfi} presents results from structured demographic evaluations on CFI-V2, designed to assess  cross-group generalization. In the age split (train on majority, test on minority age group), DCAN achieves a mean accuracy of 0.937, outperforming PMGRT (0.919) with pronounced gains in Neuroticism and Extraversion, demonstrating effective generalization to sparse behavioral patterns. In the gender split (train on mixed-gender, validate/test on single-gender subsets), DCAN attains 0.933, compared to 0.914 for PMGRT, notably narrowing the performance gap on agreeableness whcih historically sensitive to gender bias. In the ethnicity split (zero-shot transfer: train on Caucasian, test on Asian), DCAN reaches 0.932, surpassing PMGRT (0.919) with clear improvements in Openness and Conscientiousness. These results confirm that causal mechanisms enhance cross-group generalization over standard baselines.

\begin{figure}[t]
  \centering
  \begin{subfigure}[b]{0.48\linewidth}
    \centering
    \includegraphics[width=\linewidth]{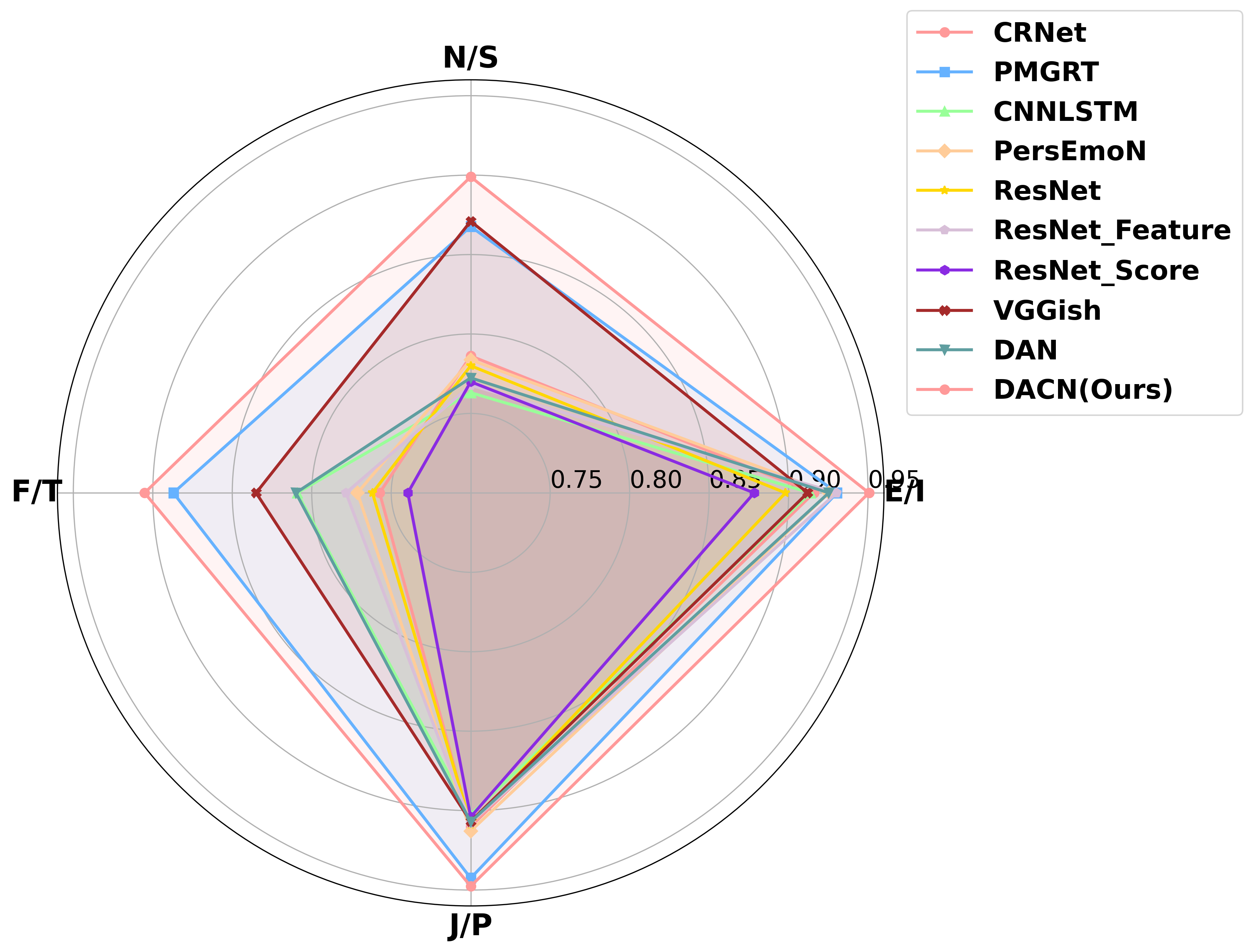}
    \caption{Age Split}
  \end{subfigure}
  \hfill
  \begin{subfigure}[b]{0.48\linewidth}
    \centering
    \includegraphics[width=\linewidth]{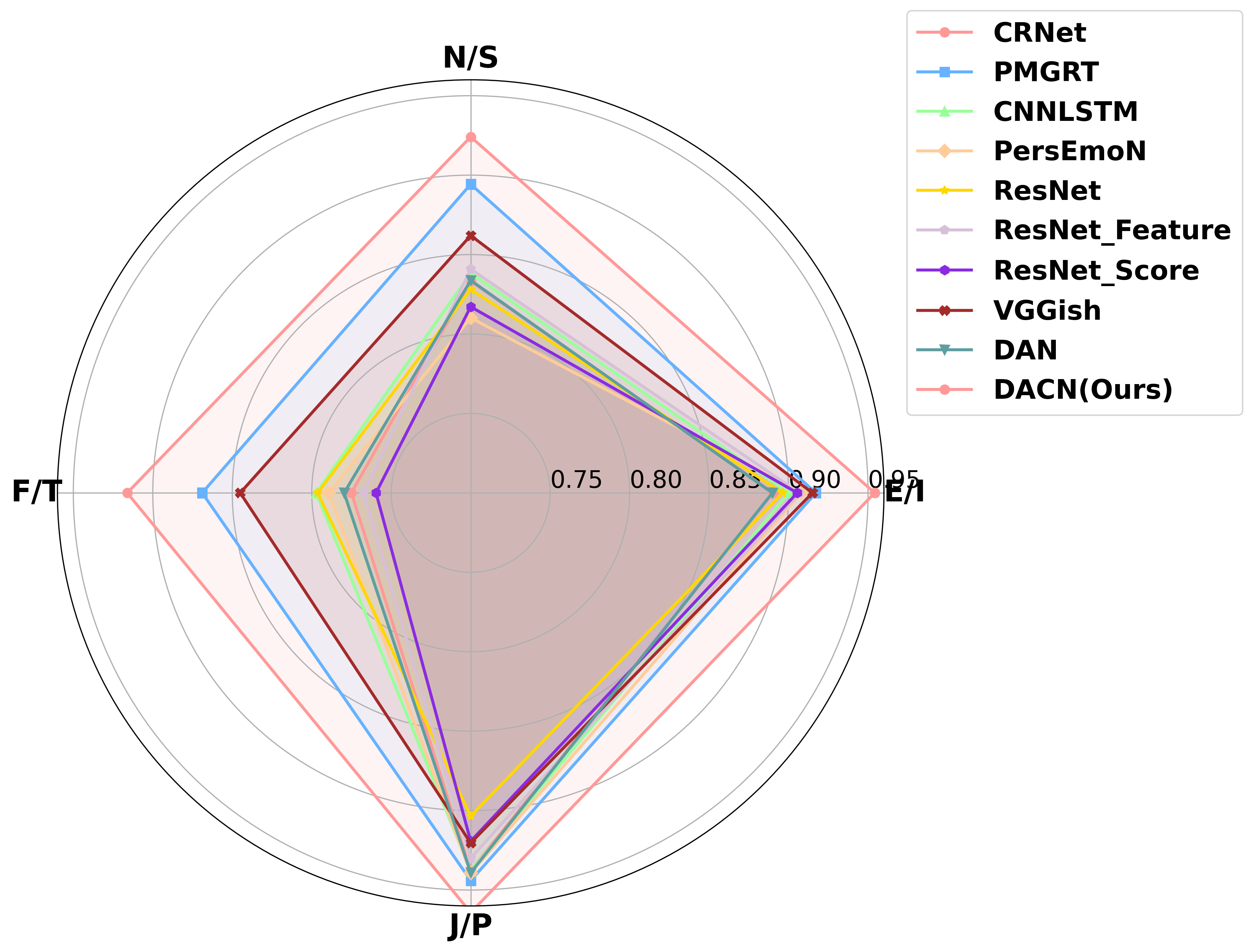}
    \caption{Gender Split}
  \end{subfigure}
  \caption{The OOD experiment on DMSP (MBTI).}
  \label{fig:demographic_radars_mbti}
\end{figure}

\textbf{(2) Out-of-Distribution on MBTI Traits.} Figure~\ref{fig:demographic_radars_mbti} evaluates DCAN on DMSP under controlled distribution shifts. In the age split, DCAN achieves 0.926 mean accuracy, compared to 0.907 for PMGRT, with consistent gains across all four dichotomies (E/I, N/S, T/F, J/P), proving resilient to age-related behavioral variations. More critically, in the gender split—trained exclusively on males and tested on females, DCAN attains 0.939, exceeding PMGRT (0.906) by a substantial margin. The largest improvement appears on the Thinking/Feeling (T/F) dimension, which is highly susceptible to gender stereotyping. This demonstrates DCAN's capability in decoupling personality traits from spurious demographic correlations, ensuring generalization under severe distribution shifts.

\subsection{Hyperparameter Analysis}
\label{app:hyperparameter}
\begin{figure}[t]
  \centering
  \begin{subfigure}[b]{0.32\linewidth}
    \centering
    \includegraphics[width=\linewidth]{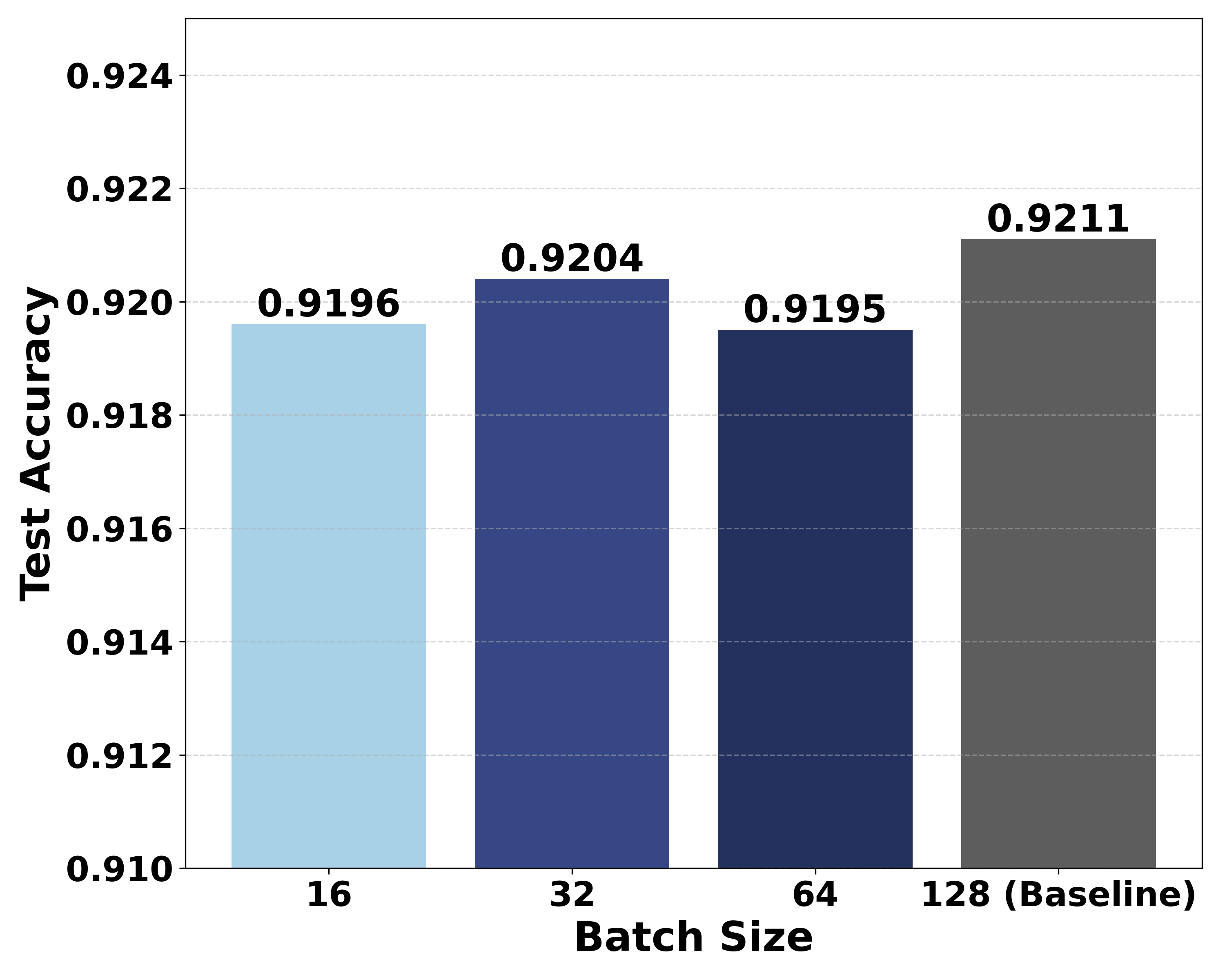}
    \caption{Batch Size }
    \label{fig:hp_bs}
  \end{subfigure}
  \hfill
  \begin{subfigure}[b]{0.32\linewidth}
    \centering
    \includegraphics[width=\linewidth]{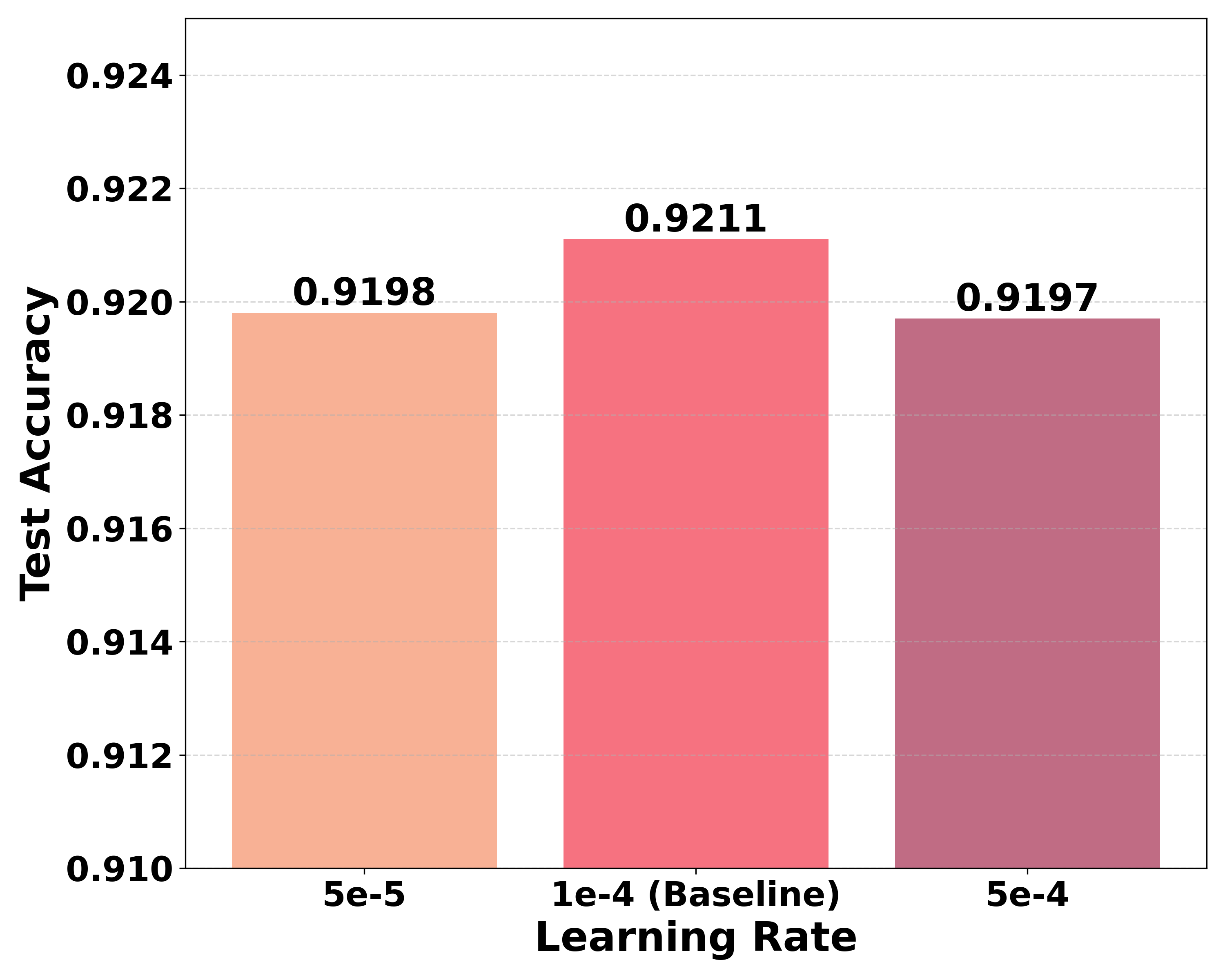}
    \caption{Learning Rate }
    \label{fig:hp_lr}
  \end{subfigure}
  \hfill
  \begin{subfigure}[b]{0.32\linewidth}
    \centering
    \includegraphics[width=\linewidth]{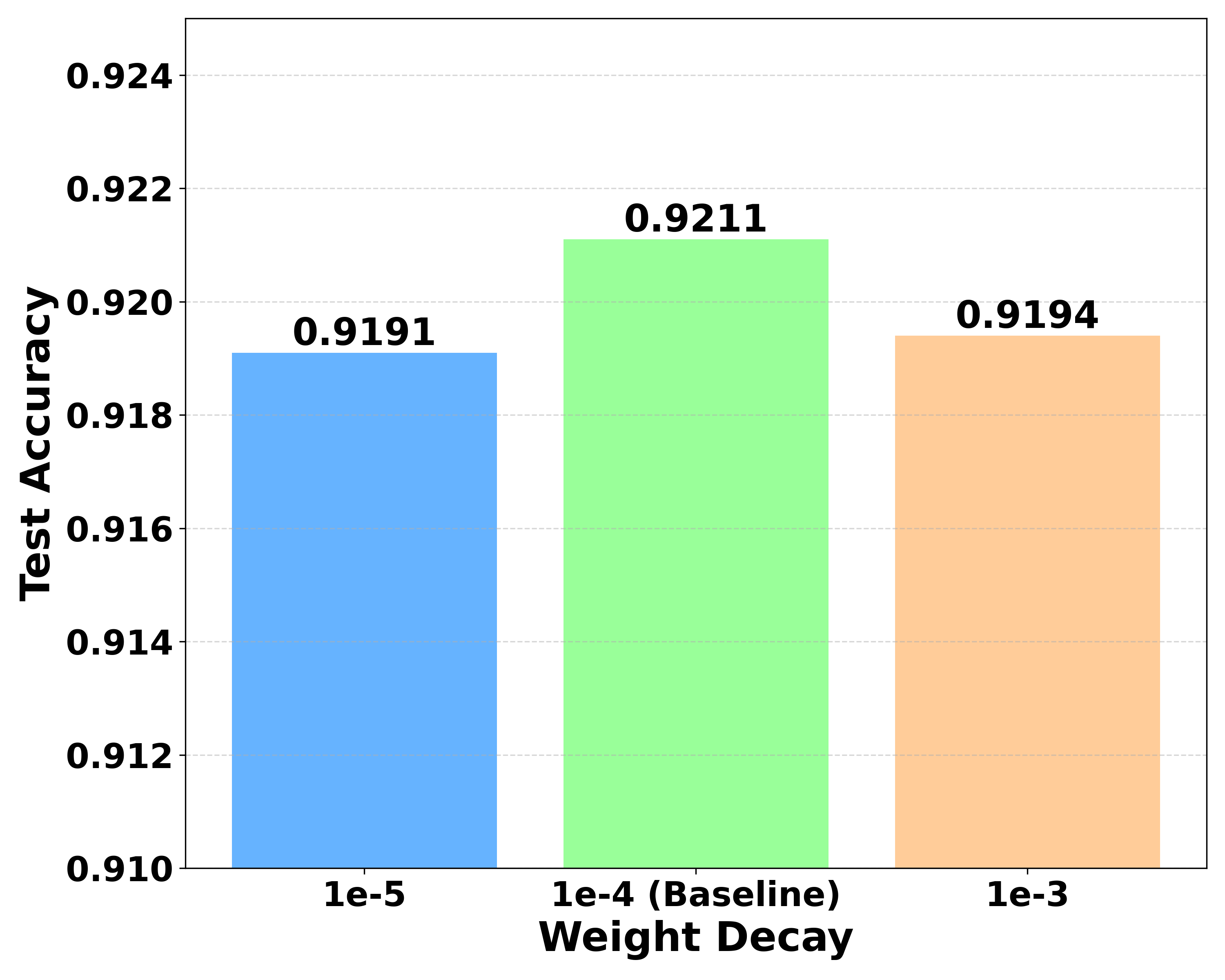}
    \caption{Weight Decay }
    \label{fig:hp_wd}
  \end{subfigure}
  \caption{Hyperparameter analysis on the CFI-V2 dataset.}
  \label{fig:hyperparameter}
\end{figure}

We conduct a sensitivity analysis on three key hyperparameters: the learning rate (LR), batch size (BS), and weight decay (WD), using the CFI-V2 dataset. As shown in Fig.~\ref{fig:hyperparameter}, DCAN exhibits stable performance across a broad range of hyperparameter settings, indicating a smooth and robust optimization landscape. Among all configurations, the optimal setting (LR=$1\times10^{-4}$, BS=32, WD=$1\times10^{-4}$) achieves the best trade-off between convergence efficiency and generalization performance. This may be because the proposed dual causal adjustment mechanism provides effective representations during training.

\subsection{Case Study}
\label{sec:case_study}
We conduct three qualitative case study demonstrate how DCAN effectively handles diverse confounding factors.

\noindent\textbf{(1) Disentangling Transient Affect.} 
The baseline model erroneously attributes common transient positive emotions (e.g., smiling) to high \textit{Openness}, confounding momentary states with stable traits. DCAN eliminates this spurious correlation by shifting focus to stable linguistic cues (e.g., prudent lexical choices), producing a prediction consistent with the ground truth.

\noindent\textbf{(2) Semantic-Visual Conflict.} 
When strong lexical cues for \textit{Extraversion} conflict with static visual scenes, the baseline underestimates the trait. DCAN aligns energetic textual content with expressive facial dynamics, successfully bridging the cross-modal gap to achieve accurate reasoning.

\noindent\textbf{(3) Identifying Trait-Diagnostic Cues.} 
When assessing \textit{Neuroticism}, the baseline overlooks subtle diagnostic signals in stressful contexts. DCAN enhances sensitivity to complex, trait-relevant patterns—such as vocal tension or micro-expressions—correctly capturing high Neuroticism where the baseline fails to detect the emotional nuance.

\begin{figure}[t]
    \centering
    \includegraphics[width=\linewidth]{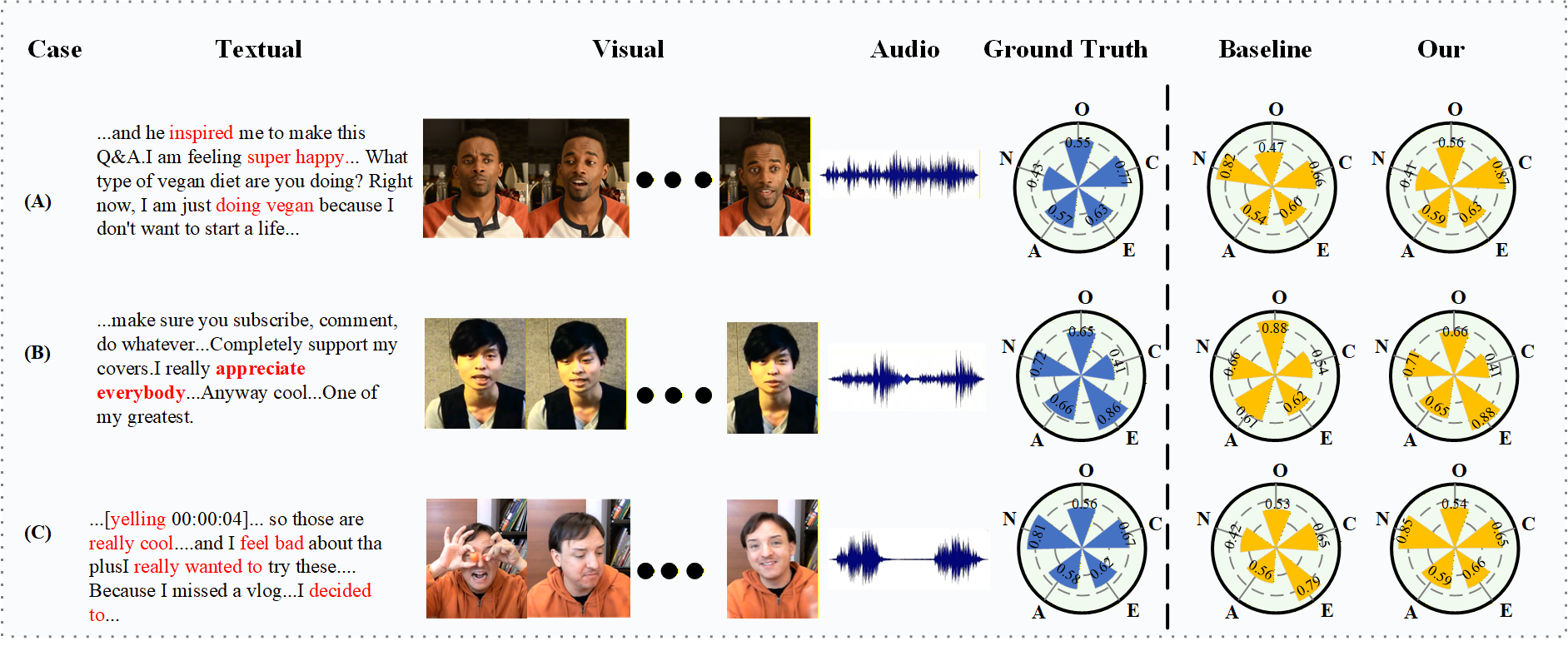}
    \caption{Qualitative comparison between the baseline and DCAN: (A) DCAN avoids openness overestimation by ignoring irrelevant smiles; (B) DCAN corrects extraversion underestimation by aligning semantic and visual cues; (C) DCAN captures neuroticism by detecting subtle stress signals.}
    \label{fig:case_study}
\end{figure}

\section{Ethical Considerations}
This study adheres to strict ethical standards with respect to privacy, transparency, and responsible use. All datasets are sourced from publicly available materials or collected with informed consent exclusively for academic research purposes. To protect participant privacy, the data are carefully anonymized to remove any personally identifiable information and are approved by an institutional review board. The primary objective of this work is to advance research in personality computing.
We explicitly caution against its deployment in high-stakes automated decision-making systems without rigorous human oversight and thorough ethical review.

\section{Conclusion}

In this paper, we investigate the causal rationale behind Individual-generated multimodal video data and traits in personality understanding. And we propose a novel dual causal adjustment network named DCAN to optimize a fair personality understanding model. 
DCAN comprises back-door and front-door adjustment causal learning, to explicitly model and eliminate both observable and latent confounders.  Besides, we construct a new Demographic-annotated Multimodal Student Personality (DMSP) dataset under the MBTI taxonomy that supports in-depth analysis of subject bias. Extensive experiments demonstrate the superiority of our DCAN in both accuracy and fairness in multimodal personality understanding.

\section*{Acknowledgements}

This work was supported by National Key Research and Development Program (No. 2024YFB4710400),National Natural Science Foundation of China (62272322, 62272323, 62403334, 62206148),Taking on Challenging Projects by Responding to Calls for Solutions Projects of Key Technological Breakthrough Projects for Major Industries in Henan Province (251000210300), the Science and Technology Breakthrough Project(No.262102211001),CIPS-SMP-Zhipu Large Model Fund (20250307),China Postdoctoral Science Foundation (2025M771594 ), and Beijing Postdoctoral Research Foundation (2025-135).

\balance
\bibliographystyle{ACM-Reference-Format}
\bibliography{sample-base}

\end{document}